\documentclass[11pt]{article}

\usepackage[preprint]{acl}
\usepackage{times}
\usepackage{latexsym}
\usepackage[T1]{fontenc}
\usepackage[utf8]{inputenc}
\DeclareUnicodeCharacter{2192}{\ensuremath{\rightarrow}}
\DeclareUnicodeCharacter{2014}{\textemdash}
\usepackage{microtype}
\usepackage{inconsolata}
\usepackage{fancyvrb}
\usepackage{graphicx}
\usepackage{booktabs}
\usepackage{tabularx}
\usepackage{array}
\usepackage{amsmath,amssymb,mathtools}
\usepackage{enumitem}
\usepackage{float}
\usepackage{placeins}
\usepackage{algorithm}
\usepackage{algpseudocode}

\graphicspath{{figures/}}

\title{SkillFocus: Evolving Agent Skills via Capability Decomposition}
\author{
  Ning Wang\textsuperscript{1} \quad
  Zhiren Gong\textsuperscript{2} \quad
  Bingdong Li\textsuperscript{1}\thanks{Corresponding author.} \quad
  Peng Yang\textsuperscript{3} \quad
  Aimin Zhou\textsuperscript{1,4} \\[2pt]
  \parbox{0.98\textwidth}{\centering\normalsize
    \textsuperscript{1}East China Normal University;
    \textsuperscript{2}Nanyang Technological University;\\
    \textsuperscript{3}Southern University of Science and Technology;
    \textsuperscript{4}Shanghai Innovation Institute
  }
}
\date{}

\begin{document}
\maketitle

\begin{abstract}
Agent skill evolution seeks to improve reusable procedural guidance for large
language model (LLM) agents through iterative revision.  Existing methods base
each revision mainly on execution trajectories or feedback, leaving recurring
behavioral requirements across tasks implicit and tying revision to the
behavior of the current skill.  We introduce SkillFocus, which decomposes
recurring task requirements into a capability space that remains fixed as the
skill evolves, separating what tasks require from how the current skill
behaves.  SkillFocus maps current task outcomes to this space to identify the
capability that leaves the most tasks unresolved, then uses that capability to
determine what to revise and which evidence to use.  Across four benchmarks
spanning heterogeneous tasks, SkillFocus achieves the best held-out accuracy on
all four, outperforming the strongest competing result by 5.7 points on average
while using 24\% fewer evolution tokens on average than the closest iterative
baseline.  Controlled studies further show that capabilities derived from recurring task
requirements outperform task-semantic and execution-derived alternatives, while
randomizing task--capability assignments reduces final accuracy by up to 20.2
points.  Matching evidence to the selected capability increases candidate gain
by 4.4 points under prioritized revision.

\end{abstract}

\section{Introduction}
\label{sec:introduction}

Large language model (LLM) agents \citep{yao2023react} can benefit from Agent
Skills, reusable procedural guidance for reasoning, tool use, and task
execution \citep{anthropic2025skills,skillsbench2026}.  Because this guidance
can be revised without updating model parameters, skill evolution seeks to
improve a skill over repeated task executions
\citep{wang2026sage,yang2026skillopt}.
When one skill is reused across heterogeneous tasks, however, its revisions
must serve tasks that require different and often overlapping procedural
requirements.  The practical challenge is therefore to improve one shared
skill across heterogeneous tasks without treating every task as a separate
revision problem.

\begin{figure}[t]
\centering
\includegraphics[width=\columnwidth]{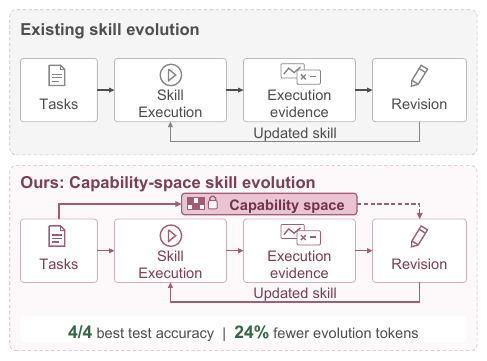}
\caption{\textbf{Existing and capability-space skill evolution.} Both follow
the same execution--revision loop.  Existing methods base revisions on current
execution evidence, whereas ours also derives a capability space from tasks
before execution and uses it during revision.}
\label{fig:concept}
\end{figure}

\begin{figure*}[!t]
\centering
\includegraphics[width=0.99\textwidth]{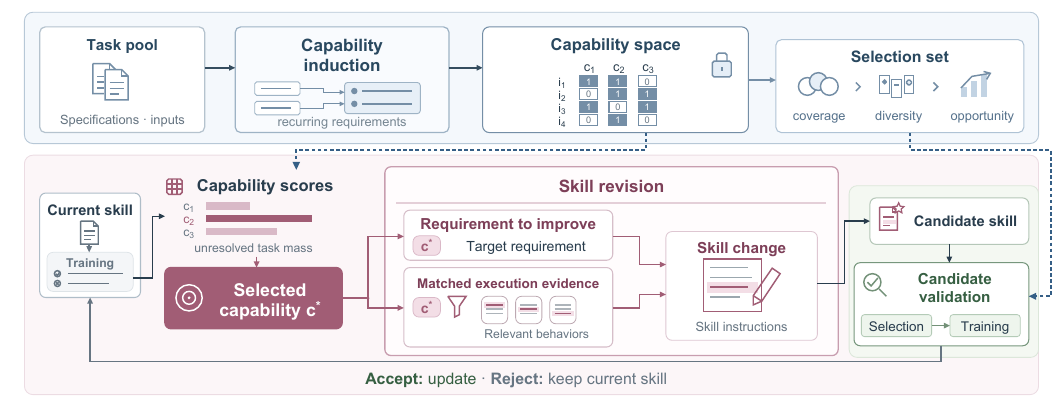}
\caption{\textbf{Overview of SkillFocus.} Before evolution, capability
induction builds a fixed capability space $(\mathcal C,Q)$ from task
specifications, from which the selection set is built. In each round,
capability scores select $c^*$, which sets the requirement to improve and the
matched execution evidence; the resulting candidate replaces the current skill
only after selection and training validation.}
\label{fig:method-overview}
\end{figure*}

Existing skill-evolution methods derive revisions mainly from execution
evidence: distilled trajectories \citep{ni2026trace2skill}, diagnosed failures
\citep{wang2026skillgrad,liu2026skillrevise}, or scored rollouts
\citep{yang2026skillopt}.  This evidence shows what the current skill did, but
it changes as the skill changes and does not reveal which procedural
requirements recur across tasks.
Using tasks themselves as revision units does not solve the problem either:
one task may require several procedures, while the same procedure may appear
in many different tasks.  Figure~\ref{fig:concept} illustrates this gap:
execution evidence remains necessary for revision, yet it cannot tell which
requirements the tasks share.

A useful unit for skill revision should be finer than a task, yet remain
meaningful as the skill changes.  Two spreadsheet tasks with different goals,
for instance, may both require checking the populated range before writing,
while one of them also requires formula construction and output verification.
We call such a recurring procedural requirement a capability when the skill
can be revised to handle it better.  Revising the skill can improve how well it
handles a requirement without changing what the task requires.  This leads to our central question: can recurring
requirements derived before execution guide both what to revise and which
execution evidence to use as the skill changes?

We introduce SkillFocus, which identifies recurring procedural requirements
from task specifications before any execution and uses them to define a
capability space that remains fixed as the skill changes
(Figure~\ref{fig:method-overview}).  After each execution round, SkillFocus
maps task outcomes to the capabilities and selects the capability with the
most unresolved supporting tasks.  The selected capability determines what
part of the skill to revise and which execution evidence to use for diagnosis
and editing.  The resulting candidate is retained only after whole-skill
validation.

Across four benchmarks spanning heterogeneous tasks, SkillFocus achieves the
best held-out accuracy on all four, outperforming the strongest competing
result by 5.7 points on average while using 24\% fewer evolution tokens on
average than the closest iterative baseline.  Controlled comparisons show that
the gain depends on the capability space itself: task-semantic and
execution-derived alternatives perform worse, and randomizing task--capability
assignments reduces final accuracy by up to 20.2 points.  Within fixed
revision states, evidence matched to the selected capability also produces
stronger candidates.  Both the capability decomposition and its use during
revision contribute to the final gains.

Our contributions are as follows:

\begin{itemize}[leftmargin=*,itemsep=1pt,topsep=2pt]
    \item We formulate skill evolution over a fixed capability space derived
    from recurring procedural requirements and reuse the same capabilities
    across revision rounds.
    \item We introduce a revision process that selects a capability from
    current task outcomes and uses it to determine both what to revise and
    which execution evidence to use.
    \item We show that SkillFocus improves final skill quality and evolution
    efficiency across heterogeneous tasks, with controlled studies supporting
    both the capability decomposition and capability-matched evidence.
\end{itemize}

\section{Related Work}
\label{sec:related-work}

\paragraph{Experience Reuse and Procedural Skills.}
Experience-driven agents reuse interaction outcomes to improve later task
execution \citep{zhang2025memorysurvey}.  Early work folds environment
interaction, memory, and tool use into the execution loop itself
\citep{yao2023react,sumers2024coala,schick2023toolformer}, while
reflection-based methods carry verbal feedback or distilled lessons into
subsequent attempts
\citep{shinn2023reflexion,madaan2023selfrefine,zhao2024expel,wu2026evolver}.
Other work keeps the same experience for later runs as executable skill
libraries \citep{wang2024voyager,wang2026sage}, reusable workflows
\citep{wang2025awm}, procedural memory \citep{mi2026procmem}, or reasoning
memories distilled from successes and failures \citep{ouyang2026reasoningbank};
memory systems likewise store and retrieve past experience on demand
\citep{xu2025amem,kang2025memoryos,yu2026agemem,xiong2026memory}.

\paragraph{Skill Evolution from Execution Evidence.}
Building on this idea, recent skill-evolution methods repeatedly propose
changes to a skill using evidence produced during task execution.  Trace2Skill
distills lessons from successful and failed trajectories into a transferable
skill \citep{ni2026trace2skill}, SkillGrad turns trajectory diagnoses into
textual update directions \citep{wang2026skillgrad}, SkillOpt revises a
retained skill from scored rollouts \citep{yang2026skillopt}, which
SkillOpt-Lite trims to a minimal loop validated on held-out samples
\citep{shen2026skilloptlite}, and SkillRevise applies trace-conditioned,
execution-anchored edits \citep{liu2026skillrevise}.  Others repair recurring
failures in batch, keep a change only after replay validation, or evolve the
agent's policy instead of a textual skill
\citep{liu2026skillforge,chen2026skillcat,he2026skillcommit,zhang2024agentpro}.
Execution evidence is useful for diagnosing current failures, but it changes
as the skill changes, whereas the procedural requirements
imposed by the tasks do not.  SkillFocus keeps these recurring requirements
fixed across rounds and uses current execution evidence to decide which
requirement to revise.

\paragraph{Text and Context Optimization.}
A broader line of work optimizes prompts, program instructions, or persistent
context in text space \citep{ramnath2025aprosurvey}.  Instruction
optimizers propose and score candidates with an LLM
\citep{yang2024opro,zhou2023ape}, search discrete tokens or follow textual
gradients \citep{shin2020autoprompt,pryzant2023protegi}, or evolve and select
populations of prompts
\citep{fernando2024promptbreeder,do2024advicl,agrawal2026gepa}; program
optimizers jointly tune instructions and demonstrations across pipeline stages
\citep{khattab2024dspy,opsahlong2024mipro,chen2024promst}; and gradient-style and context methods
propagate language feedback through text variables or update a persistent
playbook
\citep{yuksekgonul2025textgrad,zhang2026ace,ye2026mce}.  These methods study
how to improve mutable text given feedback.  The setting studied
here asks which recurring task requirement should guide each round of change
to one shared skill.

\section{Method}
\label{sec:method}

We use skill evolution for the full multi-round process of execution, revision,
validation, and update.  A skill revision is a proposed change to the current
skill within one round; it yields a candidate skill, and a skill update
replaces the current skill with a candidate that passes validation.
SkillFocus first defines the capabilities that remain fixed during evolution
(\S\ref{sec:capability-induction}), then uses the changing task outcomes to
choose a capability and propose a revision
(\S\ref{sec:capability-guided-revision}), and finally decides whether the
candidate replaces the current skill (\S\ref{sec:candidate-selection}).
Figure~\ref{fig:method-overview} summarizes the design.

\subsection{Problem Formulation}
\label{sec:problem-setting}

We study shared-skill evolution, where one mutable skill is reused across a
task pool.  Let $A$ denote the fixed non-skill components of the target agent,
including the model, tools, and execution environment.  Let $\mathcal S$ be the
space of valid skills and $S_t\in\mathcal S$ the current skill at round $t$.
Let $\mathcal D=\{\tau_i\}_{i=1}^{N}$ denote the task pool.  For each task
$\tau_i$,
$Y_i(S):=Y(\tau_i,S)\in[0,1]$ denotes its normalized evaluation score when $A$
executes the task with skill $S$, with larger values indicating better
outcomes.  Starting from an initial skill $S_0$, which may be empty, round $t$
may propose a candidate $S'_t$, and validation determines the next skill:
\begin{equation}
S_{t+1}=
\begin{cases}
S'_t, & \text{if $S'_t$ is accepted},\\
S_t, & \text{otherwise}.
\end{cases}
\label{eq:skill-update}
\end{equation}
skill evolution runs for at most $T_{\max}$ rounds and aims to improve task
performance.
Across rounds, the skill may change, while the task pool and all non-skill
components of the agent remain fixed.

\subsection{Capability Space}
\label{sec:capability-induction}

Before evolution, SkillFocus identifies procedural requirements from each
task's specification, input structure, and environment contract, without using
any execution outcome.  We call a recurring procedural requirement a
capability when revising the skill can improve how well it is handled.
Requirements from different tasks are merged when the same reusable skill
guidance can address them.  Tasks that look similar need not share the same
procedural requirement, whereas tasks assigned to the same capability do so by
construction.  Both conditions are needed.  Reading
requirements from task specifications prevents the capabilities from changing
with the current skill, while requiring shared skill guidance means that
revising the skill can improve how those requirements are handled.

Let $\mathcal C=\{c_d\}_{d=1}^{K}$ be the induced capabilities and
$Q\in\{0,1\}^{N\times K}$ the task--capability Matrix
\citep{junker2001cognitive,wang2020neuralcd}, with
\begin{equation}
Q_{i,d}=
\begin{cases}
1, & \tau_i \text{ requires } c_d,\\
0, & \text{otherwise}.
\end{cases}
\label{eq:matrix}
\end{equation}
The entries record requirements, not outcomes.  A task may require several
capabilities, and one capability may be shared by many tasks.  We call the pair
$(\mathcal C,Q)$ the \emph{capability space}.  The tasks define the
requirements; revising the skill changes how well they are handled.  We fix
$\mathcal C$ and $Q$ before evolution begins: task outcomes may change after
each accepted skill update, while $\mathcal C$ and $Q$ do not.  Keeping the
same capabilities across rounds lets successive skills be compared against the
same task requirements.  Appendix~\ref{app:pseudocode} details the extraction,
consolidation, assignment, and support filtering procedures.

\subsection{Capability-Guided Revision}
\label{sec:capability-guided-revision}

Before the first round, the task pool is split into a training set
$D_{\rm tr}$, whose executions supply revision evidence, and a selection set
$D_{\rm sel}$, which during evolution is used only to check candidates
(\S\ref{sec:candidate-selection}).  In each round, SkillFocus uses current
training outcomes to choose one capability, and then uses that capability to
choose the evidence for revising the skill.

\paragraph{Choosing the Next Capability.}
Let $O_t$ be the execution record of $S_t$ on $D_{\rm tr}$, and write
$y_i^t:=Y_i(S_t)$.  Each capability has a support share
$\pi_d=|\mathcal D|^{-1}\sum_{\tau_i\in\mathcal D}Q_{i,d}$, fixed with the
capability space,
and a current score on the training tasks that require it:
\begin{equation}
q_d^t=\frac{\sum_{\tau_i\in D_{\rm tr}}Q_{i,d}\,y_i^t}
           {\sum_{\tau_i\in D_{\rm tr}}Q_{i,d}}.
\label{eq:capability-state}
\end{equation}
We collect these as the capability scores $\mathbf q^t=(q_d^t)_{d=1}^{K}$.
Neither coverage nor current score is sufficient alone.  A rare capability may
be poorly handled but affect few tasks, while a common capability may already
be well handled.  SkillFocus therefore prioritizes their product, the
unresolved task mass
\begin{equation}
m_d^t=\pi_d\left(1-q_d^t\right),
\label{eq:profile-priority}
\end{equation}
and selects
\begin{equation}
d_t^\star=\arg\max_{d:\,c_d\in\mathcal U_t}m_d^t,\qquad c_t^\star=c_{d_t^\star}.
\label{eq:selected-capability}
\end{equation}
The eligible capability set $\mathcal U_t$ contains the capabilities that a
round can both act on and check: each has at least one diagnosable failure in
$D_{\rm tr}$, meaning a training execution that fails for a task-logic reason
and not an environment fault (Appendix~\ref{app:selection-construction}), and
at least one supporting task in $D_{\rm sel}$.

\paragraph{Matching Evidence to the Capability.}
The selected capability decides which execution evidence the revision sees.  A
trace records everything the agent did, and most of it is unrelated to any
single requirement.  Let $\mathcal H_t$ contain the revision attempts and
outcomes from rounds before $t$.  SkillFocus keeps only the training tasks with
$Q_{i,d_t^\star}=1$ and assembles an evidence bundle $\mathcal E_t$ from them:
executions that show how the requirement was handled, the skill content
related to $c_t^\star$, and the entries of $\mathcal H_t$ associated with
$c_t^\star$.  Successful executions are set beside the failures whenever the
record contains any.
The round thus diagnoses one requirement from the executions that bear on it,
instead of from all executions pooled together.

\paragraph{Proposing the Revision.}
Diagnosis turns this evidence into a skill revision.  SkillFocus proposes a
revision only when the evidence links the selected requirement, the observed
executions, and editable skill content.  The editor then adds, replaces, or
removes skill content.  Otherwise, SkillFocus abstains on that capability and
considers the next one in $\mathcal U_t$ within the same round.  Edits that
copy task-specific entities are discarded before execution.

\begin{algorithm}[!ht]
\caption{SkillFocus}
\label{alg:skillfocus}
\small
\begin{algorithmic}[1]
\Require Fixed agent components $A$, initial skill $S_0$, task pool $\mathcal D$,
         selection size $M$, horizon $T_{\max}$
\Ensure Final skill $S$
\State Induce $(\mathcal C,Q)$ from task specifications, inputs, and
       environment contracts
\State $S\gets S_0$, $\mathcal H\gets\emptyset$
\State Execute $A$ with $S$ on $\mathcal D$ to obtain $O_0$
\State $(D_{\rm sel},D_{\rm tr})\gets
       \Call{ConstructSelection}{\mathcal C,Q,O_0,M}$
\State $O\gets O_0$ restricted to $D_{\rm tr}$
\State Compute capability scores $\mathbf q$ from $O$ and $Q$
\For{$t=0$ to $T_{\max}-1$}
    \State $\mathcal U\gets\mathcal U_t$ from $\mathbf q$
    \Repeat
        \If{$\mathcal U=\emptyset$}
            \State \Return $S$
        \EndIf
        \State $d_t^\star\gets\arg\max_{d:\,c_d\in\mathcal U}m_d^t$;\ \
               $\mathcal U\gets\mathcal U\setminus\{c_{d_t^\star}\}$
        \State Collect evidence $\mathcal E_t$ for $c_t^\star$ from $O$, $S$,
               and $\mathcal H$
        \State Diagnose, then propose $S'_t$ or abstain
    \Until{a candidate $S'_t$ is proposed}
    \State Evaluate $S'_t$ against $S$ on $D_{\rm sel}$, then on $D_{\rm tr}$
    \If{both paired changes are positive}
        \State $S\gets S'_t$; set $O$ to the training executions of $S'_t$
               from validation; recompute $\mathbf q$
    \EndIf
    \State Append the revision records and outcome to $\mathcal H$
\EndFor
\State \Return $S$
\end{algorithmic}
\end{algorithm}

\subsection{Candidate Validation}
\label{sec:candidate-selection}

\paragraph{Selection-Set Construction.}
The selection set is built once, before the first round, by greedy selection
over the task pool.  Its priorities are ordered: capability coverage first,
measured by the same support share $\pi_d$ that weights unresolved task mass
in Equation~\ref{eq:profile-priority}; then diversity over predefined task
features, namely task family, input structure, and environment contract
labels; then improvement opportunity, measured by the share of a task's
initial executions that fail for diagnosable reasons.  A reserve of tasks
solved by $S_0$ keeps regressions visible, and at least one supporting task
for every capability remains in the training set, so revision evidence remains
available for that capability.  The remaining tasks form the training set.
Both sets remain fixed after construction.  The selection set determines which
capabilities have checking tasks and so can enter $\mathcal U_t$, while
the capability selected in a later round changes neither set.
Algorithm~\ref{alg:construct-selection} gives the procedure.

\paragraph{Whole-Skill Validation.}
Because capabilities overlap and each revision edits the shared skill, a
revision aimed at one capability can change outcomes on tasks outside it.
Checking the candidate only on the tasks of that capability could accept
revisions that repair one requirement while degrading others.  SkillFocus
evaluates every candidate as a whole instead.  For
$X\in\{D_{\rm sel},D_{\rm tr}\}$, the paired change is
\begin{equation}
\begin{aligned}
\Delta_X(S',S)&:=\frac{1}{|X|}\sum_{\tau_i\in X}\\
&\quad\left[Y_i(S')-Y_i(S)\right].
\end{aligned}
\label{eq:paired-gain}
\end{equation}
A candidate is retained only when both paired changes are positive:
\begin{equation}
\begin{aligned}
\textsc{Accept}(S')
&\Longleftrightarrow
\Delta_{D_{\rm sel}}(S',S)>0\\
&\quad\land\ \Delta_{D_{\rm tr}}(S',S)>0.
\end{aligned}
\label{eq:acceptance-rule}
\end{equation}
The two checks serve different roles.  selection execution traces are not
supplied to candidate generation, so the first check evaluates the candidate
on tasks whose execution traces did not inform that revision.  The training re-evaluation checks that the
candidate produces a positive net change over the training set, which supplies
revision evidence across rounds, and it provides the execution record for the
next round.

If the candidate is accepted, $S_{t+1}=S'_t$, and its training executions from
validation become $O_{t+1}$, from which the next capability scores
$\mathbf q^{t+1}$ are computed.  Otherwise, $S_{t+1}=S_t$ and $O_{t+1}=O_t$.
All capability attempts made in round $t$, together with the proposed
revision and its validation outcome when a candidate is produced, are appended
to $\mathcal H_{t+1}$.  A round ends once a candidate reaches
validation, and the loop returns the final skill after the evolution horizon,
or earlier when no capability in $\mathcal U_t$ yields a candidate.
Algorithm~\ref{alg:skillfocus} summarizes the loop.

\section{Experiments}
\label{sec:experiments}

\begin{table*}[!t]
\centering
\caption{\textbf{Test accuracy (\%) of SkillFocus and baselines on four benchmarks.}
Bold and underlined values denote the best and second-best results,
respectively; Avg.: unweighted mean over the four benchmarks.}
\label{tab:main-results}
\small
\setlength{\tabcolsep}{5pt}
\begin{tabular}{@{}lrrrrr@{}}
\toprule
Method & SearchQA & SpreadsheetBench & LiveMath & IFBench & Avg. \\
\midrule
No Skill & 64.5 & 21.8 & 14.4 & 67.8 & 42.1 \\
One-shot Skill & 70.7 & 34.0 & 30.6 & \underline{80.6} & 54.0 \\
Trace2Skill & 73.7 & 30.7 & 39.5 & 76.3 & 55.1 \\
GEPA & 74.6 & 58.9 & 37.9 & 73.4 & 61.2 \\
SkillOpt & \underline{76.1} & \underline{62.1} & \underline{41.1} & 73.9 &
\underline{63.3} \\
\textbf{SkillFocus} & \textbf{80.1} & \textbf{77.1} & \textbf{43.6} &
\textbf{82.0} & \textbf{70.7} \\
\bottomrule
\end{tabular}
\end{table*}

\begin{figure*}[!t]
\centering
\includegraphics[width=\textwidth]{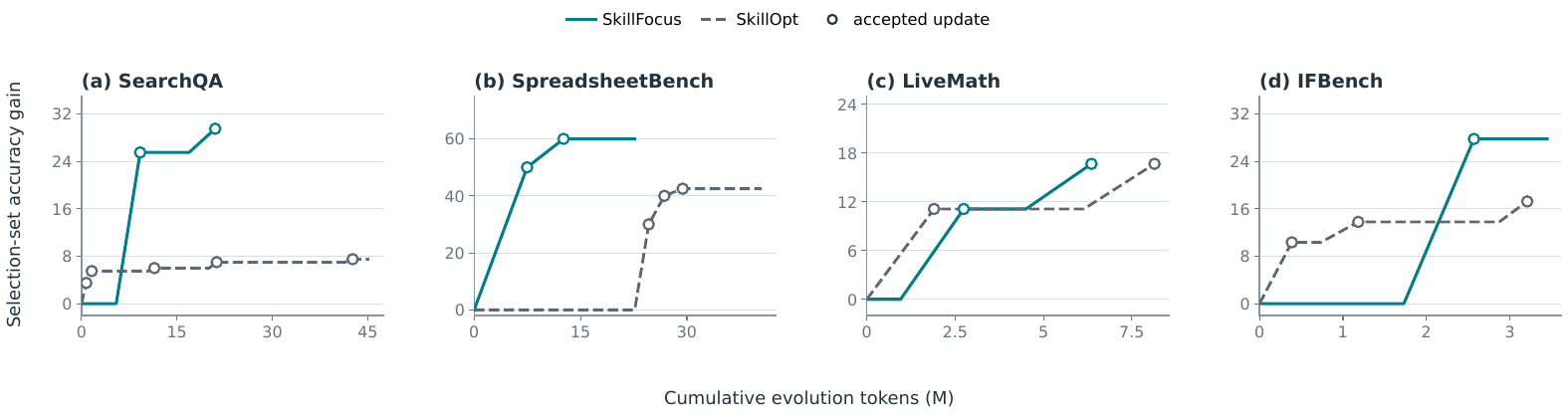}
\caption{\textbf{Evolution trajectories of SkillFocus and SkillOpt.} Each curve
plots selection set accuracy gain over the initial skill in percentage points;
markers show accepted updates, and horizontal segments are rejected candidates
that consume tokens without changing the retained skill.}
\label{fig:evolution-trajectories}
\end{figure*}

We organize our evaluation around three research questions (RQs):
\begin{itemize}[leftmargin=*,itemsep=1pt,topsep=2pt]
    \item \textbf{Research Question 1 (RQ1):} Does SkillFocus improve final
    skill quality and evolution efficiency? (\S\ref{sec:overall-results})
    \item \textbf{Research Question 2 (RQ2):} Does deriving capabilities from
    recurring task requirements improve skill evolution compared with
    task-semantic and execution-derived alternatives?
    (\S\ref{sec:capability-representation})
    \item \textbf{Research Question 3 (RQ3):} Does selecting what to revise and
    which evidence to use with the same capability improve candidate quality?
    (\S\ref{sec:coupling-results})
\end{itemize}

\subsection{Experimental Setup}
\label{sec:experimental-setup}

\paragraph{Benchmarks.}
We evaluate SkillFocus on four heterogeneous benchmarks.  SearchQA provides
search-oriented question answering tasks \citep{dunn2017searchqa}.
SpreadsheetBench (SSB) evaluates manipulation of real-world workbooks
\citep{ma2024spreadsheetbench}.  LiveMathematicianBench (LiveMath) poses
research-level mathematical reasoning as multiple-choice questions drawn from
recent papers \citep{he2026livemath}.
IFBench measures generalization to verifiable instruction constraints
\citep{pyatkin2025ifbench}, extending the verifiable constraints of IFEval
\citep{zhou2023ifeval}.  Each benchmark evolves its own shared skill, testing
the same method across distinct task types.

\paragraph{Baselines.}
We compare against five baselines.  No Skill runs the target model without
procedural guidance; One-shot Skill writes a skill once without execution
feedback; Trace2Skill \citep{ni2026trace2skill} distills a skill from execution
trajectories; Genetic-Pareto (GEPA) \citep{agrawal2026gepa} mutates the skill
text through reflection and keeps candidates on a Pareto front; and SkillOpt
\citep{yang2026skillopt} iteratively revises a retained skill from rollout
evidence.

\paragraph{Protocol.}
Within each benchmark, all methods use the same evaluator, task pool,
held-out test split, and DeepSeek-V4-Flash target model
\citep{deepseek2026v4}; the skill-optimization methods also share the initial
skill and use the Generative Pre-trained Transformer model GPT-5.4
\citep{openai2026gpt54} for optimizer-side calls.  Each method follows its own
candidate-selection procedure within the same non-test pool, and no method
receives additional tasks.

\paragraph{Metrics.}
We report held-out test accuracy and cumulative evolution tokens, measuring
final skill quality and optimization cost, respectively.  Evolution cost counts
all inference tokens spent before the final skill is fixed; for SkillFocus,
this includes capability induction and the initial executions used before
iterative revision.  The final test evaluation is not counted.
Appendix~\ref{app:evaluation-details} gives the full protocol, including data
splits, optimization budgets, execution settings, and cost accounting.

\subsection{Main Results}
\label{sec:overall-results}

SkillFocus achieves the highest held-out test accuracy on all four benchmarks
(Table~\ref{tab:main-results}).  Its margin over the strongest competing
result on each benchmark ranges from 1.4 to 15.0 points and averages 5.7
points, with the widest margin over SkillOpt, the closest iterative baseline,
on SpreadsheetBench.  On IFBench, SkillFocus is the only iterative method that
exceeds One-shot Skill, reaching 82.0 against 80.6.
Sections~\ref{sec:capability-representation} and~\ref{sec:coupling-results}
examine which parts of capability-guided revision produce this advantage.

SkillFocus reaches these accuracies while using 24\% fewer evolution tokens
than SkillOpt on average, with the largest reduction on SearchQA (54.3\%).
Figure~\ref{fig:evolution-trajectories} shows that it reaches larger gains
through fewer accepted revisions and, on the first three benchmarks, makes its
last accepted revision earlier in the token budget.  This is consistent with
prioritizing the capability that leaves the most tasks unresolved, so one
accepted revision can move many tasks at once.  Replacing the constructed selection set with a
uniform random sample of the same size also lowers final test accuracy by
3.9--17.2 points (Appendix~\ref{app:analysis-efficiency}).  Taken together,
SkillFocus produces a better final skill on every benchmark while using fewer
evolution tokens than SkillOpt on average.

\begin{figure*}[!t]
\centering
\includegraphics[width=0.98\textwidth]{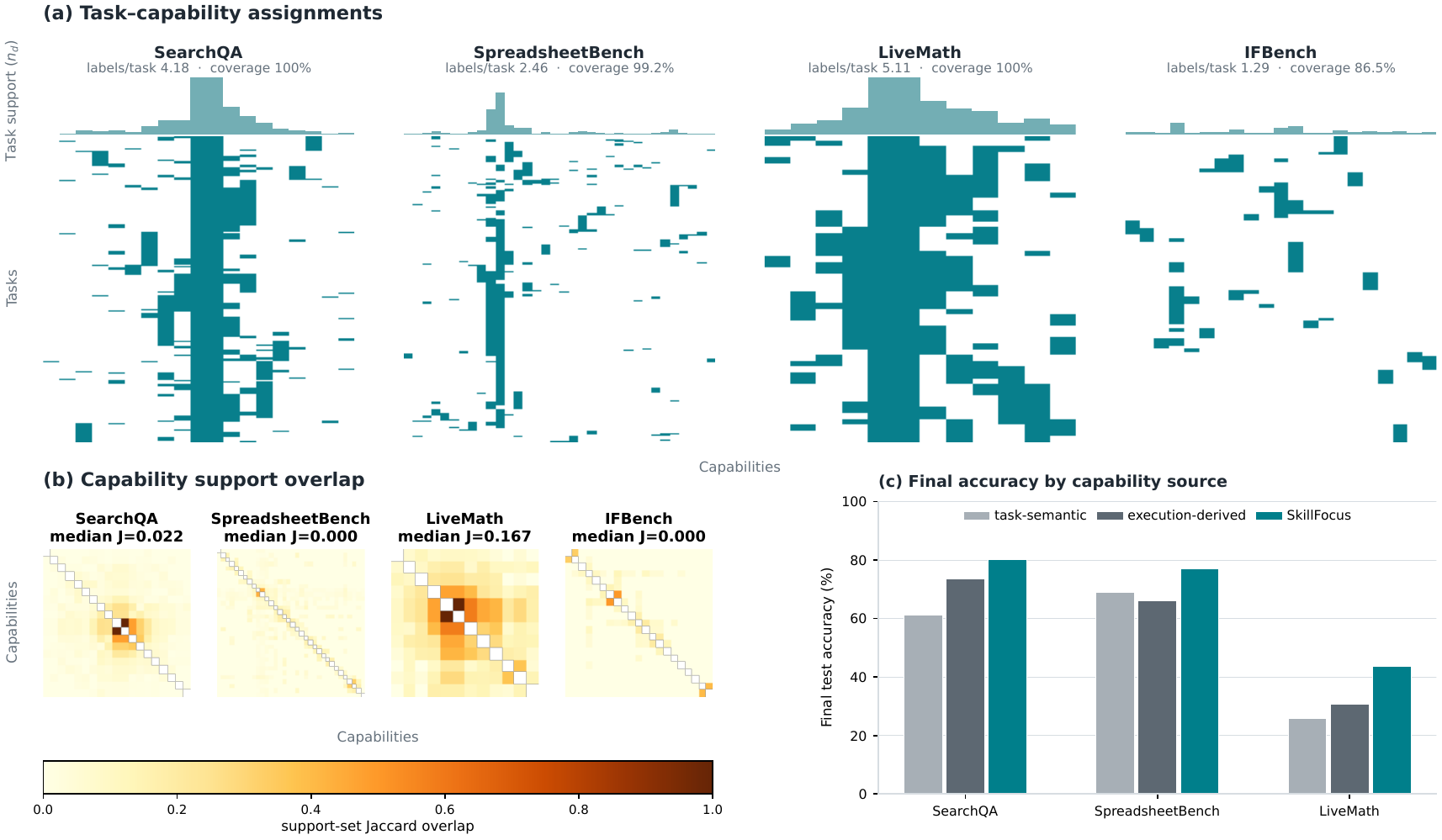}
\caption{\textbf{Task--capability assignments and their effect on final skill
accuracy.} (a) capabilities assigned to each task and the number of tasks per
capability; (b) overlap between the task sets of different capabilities;
(c) final skill accuracy when capabilities are derived from different
sources. Panel annotations report labels per task, coverage, and median
support overlap.}
\label{fig:capability-structure}
\end{figure*}

\subsection{Analysis of the Capability Space}
\label{sec:capability-representation}

Tasks often require several capabilities, while different capabilities cover
different subsets of tasks.  On SearchQA, SpreadsheetBench, and LiveMath, at
least 86\% of tasks require multiple capabilities, and 36\% do so on IFBench
(Figure~\ref{fig:capability-structure}a).  The median Jaccard overlap between
the task sets of two capabilities stays at or below 0.17 on every benchmark
(Figure~\ref{fig:capability-structure}b).

\emph{Where capabilities come from} changes the quality of the evolved skill.
Task-semantic grouping uses overall task similarity, whereas execution-derived
grouping uses the outcomes of the initial skill.  With the rest of the pipeline
fixed, capabilities derived from procedural requirements outperform the former
by 8.2--18.8 points and the latter by 6.6--12.9 points on all three evaluated
benchmarks (Figure~\ref{fig:capability-structure}c), and which alternative
comes second varies by benchmark.  Unlike the execution-derived grouping, these
requirements are defined by the tasks and stay the same as the skill
changes.

Which tasks share a capability matters.  The randomized control preserves how
many capabilities each task has and how many tasks each capability covers, but
breaks which tasks are linked to which requirements, and final accuracy drops
by 11.1--20.2 points (Table~\ref{tab:representation-endpoints}).  The gain
therefore comes from the exact task--capability assignments, not only from the
number or size of the capability groups.

Independent proposal views also recover similar task--capability assignments
on all four benchmarks, with an adjusted Rand index (ARI) of 0.883--0.973 and
normalized mutual information (NMI) of 0.953--0.989
(Appendix~\ref{app:stability}).  Together, these results show that deriving
capabilities from recurring task requirements yields better final skills than
the task-semantic and execution-derived alternatives, and that the specific
task--capability assignments are both consequential and reproducible.

\subsection{Analysis of Capability-Guided Revision}
\label{sec:coupling-results}

We next test whether candidates improve when the revision target and its
evidence refer to the same capability.  We hold 15 historical revision states
fixed and rerun candidate generation under a $2\times2$ crossover: the
selected capability is either the highest-priority or a random non-top
capability, and the evidence is either matched or mismatched to that
capability (Table~\ref{tab:focus-evidence-crossover}).  Each condition thus
starts from the same current skill and execution history; only the chosen
capability and the evidence supplied to revision change.

\begin{table}[t]
\centering
\caption{\textbf{Focus--evidence} crossover over 15 historical revision states,
macro-averaged across SearchQA, SpreadsheetBench, and LiveMath. Accept:
candidate acceptance rate; $\Delta_{\rm sel}$, $\Delta_{\rm tr}$: paired
candidate gains (points) on the selection and training sets.}
\label{tab:focus-evidence-crossover}
\small
\setlength{\tabcolsep}{3pt}
\begin{tabular}{@{}lrrr@{}}
\toprule
Condition & Accept (\%) & $\Delta_{\rm sel}$ & $\Delta_{\rm tr}$ \\
\midrule
\textbf{TM: Top + Matched} & \textbf{40.0} & \textbf{+4.10} & \textbf{+9.37} \\
T$\neg$M: Top + Mismatched & 26.7 & +3.62 & +4.97 \\
RM: Random + Matched & 26.7 & +2.28 & +5.33 \\
R$\neg$M: Random + Mismatched & 20.0 & +2.09 & +3.39 \\
\bottomrule
\end{tabular}
\end{table}

Candidates are strongest when SkillFocus revises the highest-priority
capability using evidence from tasks that require that same capability.  We
compare candidates by the training gain $\Delta_{\rm tr}$ at the same
historical state, since the training executions supply the revision evidence.
Matched evidence raises $\Delta_{\rm tr}$ for both the highest-priority and a
random capability, and top-priority selection raises it under both matched and
mismatched evidence.  Under top-priority selection, matched evidence adds 4.4
points; Top + Matched reaches the highest $\Delta_{\rm tr}$ ($+9.37$) and the
highest acceptance rate (40.0\%), so stronger candidates also pass validation
more often.

Whole-run ablations show the same pattern: removing capability-matched
evidence, or removing both matched evidence and capability prioritization,
lowers final skill accuracy on all three evaluated benchmarks
(Appendix~\ref{app:coupling-analysis}).  The crossover isolates candidate
quality from the same starting state, while the whole-run ablations show that
these per-round gains reach the final skill.  Using one capability to choose
both what to revise and which evidence to read therefore improves the
candidates a round produces, and those gains carry through to the final
skill.

\FloatBarrier
\section{Conclusion}
\label{sec:conclusion}

We presented SkillFocus, a framework for evolving a shared Agent skill through
a fixed capability space derived from recurring procedural requirements.  As
the skill changes, SkillFocus uses current task outcomes to choose which
capability to improve, and uses that same capability to select the execution
evidence for revision.  Across heterogeneous benchmarks, SkillFocus produces
stronger final skills at a lower average evolution cost, and controlled studies
support both the capability space and its use to guide revision.  More broadly,
skill evolution can separate what tasks repeatedly require from how the current
skill behaves, keeping the former fixed and using the latter to decide what to
improve and which evidence to revise from.

\section*{Limitations}

SkillFocus assumes that recurring procedural requirements can be identified
before evolution and that these requirements are addressable through skill
revision.  Because each revision changes the shared skill as a whole, improving
one capability can also change behavior on other requirements; SkillFocus
validates every candidate as a whole rather than confining an edit to its
target.  Our evaluation studies bounded evolution under one optimizer--target
configuration; broader model families and longer-horizon or cross-domain
evolution remain future directions.

\bibliography{references}

\clearpage
\appendix
\section*{Appendix Overview}

Appendix~\ref{app:additional-analysis} expands the evidence for the two central
claims of the paper: the capability space groups tasks in a structured and
reproducible way that changes the evolved skill, and capability-guided
revision improves candidate quality.  Appendix~\ref{app:evaluation-details} documents the
evaluation protocol, fairness controls, and cost accounting.
Appendix~\ref{app:method-details} details each stage of the method with its
pseudocode, Appendix~\ref{app:prompt-templates} gives representative prompt
interfaces, and Appendix~\ref{app:llm-usage} states how large language models
are used.

\section{Extended Analysis}
\label{app:additional-analysis}

Beyond the two central claims, this section examines how far a whole-skill
revision reaches beyond its target tasks, how selection accuracy relates to
held-out accuracy, where the evolution budget goes, and one complete revision
chain.

\subsection{Capability Structure and Functional Relevance}
\label{app:analysis-capability}

No fixed human taxonomy covers the induced capabilities, so we evaluate them
by their structure, their stability across proposal views, and the skill they
produce, instead of by a ground-truth clustering score.

\paragraph{Structure.}
A task usually requires several capabilities, and each capability covers a
different number of tasks.  The mean number of active capabilities per task
ranges from 1.29 on IFBench to 5.11 on LiveMath, while support ranges from
small task subsets to requirements shared by the whole pool
(Table~\ref{tab:capability-summary}).  Figure~\ref{fig:capability-structure}
shows the same three facts as panels: which capabilities each task carries,
how far their task sets overlap, and what the resulting skill scores.

\begin{table*}[t]
\centering
\caption{Benchmark-level statistics of the frozen task--capability matrices.
Tasks is the number of task records the matrix covers; $K$ is the number
of induced capabilities; Labels per task is the mean number of active
capabilities per task; Support is the minimum--maximum number of supporting
tasks; $J$ is pairwise support-set Jaccard overlap; and Coverage is the
fraction of covered tasks carrying at least one capability.}
\label{tab:capability-summary}
\small
\setlength{\tabcolsep}{3pt}
\begin{tabular}{@{}lrrrrrrr@{}}
\toprule
Benchmark & Tasks & $K$ & Labels per task & Support & Mean $J$ & Median $J$ & Coverage \\
\midrule
SearchQA & 600 & 19 & 4.18 & 3--600 & .055 & .022 & 100\% \\
SpreadsheetBench & 120 & 34 & 2.46 & 3--89 & .018 & .000 & 99.2\% \\
LiveMath & 53 & 12 & 5.11 & 5--53 & .202 & .167 & 100\% \\
IFBench & 89 & 21 & 1.29 & 3--18 & .018 & .000 & 86.5\% \\
\bottomrule
\end{tabular}
\end{table*}

\begin{table*}[t]
\centering
\caption{Representative behavioral requirements and their task support.}
\label{tab:capability-examples}
\small
\setlength{\tabcolsep}{3pt}
\begin{tabularx}{\textwidth}{@{}lXr@{}}
\toprule
Benchmark & Representative behavioral requirement & \# Tasks \\
\midrule
SearchQA & Return exactly one non-empty answer element containing only the final answer text, with no reasoning, citations, or extra prose. & 600 \\
SearchQA & Use only the supplied question, documents, and metadata as evidence, avoiding outside knowledge, tools, and unsupported inference. & 600 \\
SearchQA & Select an answer only when the same candidate satisfies all substantive clue constraints, rather than combining separate partial matches. & 191 \\
SearchQA & Follow the relation expressed in the clue to extract the correct participant, such as an author, actor, founder, owner, or creator. & 127 \\
SpreadsheetBench & Write the exact formula or formula-based expression required by the task directly in the specified answer cells. & 89 \\
SpreadsheetBench & Generate each output with correct row-specific references, conditions, and source matches rather than positional assumptions. & 31 \\
SpreadsheetBench & Identify and process rows by comparing the required keys, text, or filter conditions so only matching records contribute. & 17 \\
LiveMathematicianBench & Base the answer only on the supplied prompt and choices, avoiding outside knowledge, retrieval, tools, and code execution. & 53 \\
LiveMathematicianBench & Select the strongest claim supported by the prompt while rejecting weaker consequences and unsupported overstatements. & 31 \\
LiveMathematicianBench & Preserve exact quantifier order, object scope, endpoint cases, and domain restrictions when comparing choices. & 22 \\
IFBench & Use the structural conventions of the requested output genre, document form, or presentation format rather than generic prose. & 18 \\
IFBench & Satisfy exact, minimum, maximum, distinctness, or repetition counts for specified lexical categories while avoiding unintended extras. & 13 \\
IFBench & End every sentence or sentence-like line with an emoji when the instruction requires it. & 3 \\
\bottomrule
\end{tabularx}
\end{table*}

Table~\ref{tab:capability-examples} gives representative
behaviors grouped into capabilities. SpreadsheetBench combines a high-support formula-writing
behavior with narrower matching and reference behaviors, while IFBench
contains sparse requirements attached to smaller task subsets. A task may
carry several capabilities at once.

\paragraph{Functional relevance.}
\begin{table}[!ht]
\centering
\caption{Test accuracy (\%) when capabilities are derived from different
sources, under the same downstream SkillFocus pipeline.}
\label{tab:representation-endpoints}
\small
\setlength{\tabcolsep}{3pt}
\begin{tabular}{@{}lrrr@{}}
\toprule
capability source & SearchQA & SSB & LiveMath \\
\midrule
\textbf{SkillFocus} & \textbf{80.1} & \textbf{77.1} & \textbf{43.6} \\
task-semantic & 61.3 & 68.9 & 25.8 \\
execution-derived & 73.5 & 66.1 & 30.7 \\
randomized assignment & 69.0 & 59.6 & 23.4 \\
\bottomrule
\end{tabular}
\end{table}
Table~\ref{tab:representation-endpoints} tests three alternative explanations
for the benefit of the capability space.  Each alternative replaces one
property of the frozen $(\mathcal C,Q)$ while the downstream evolution
pipeline stays fixed (Appendix~\ref{app:decomposition-source}).

\emph{Not generic task grouping.}  The task-semantic decomposition groups tasks by
their overall objective, operation, and domain, reading the same task
specifications and allowed the same number of capabilities.  It trails
SkillFocus by 18.8, 8.2, and 17.8 points on SearchQA, SpreadsheetBench, and
LiveMath.  Tasks with similar surface semantics may share no behavior that one
skill edit can change, and a task-semantic label keeps a task whole instead of
splitting it into the separate requirements a revision can address.

\emph{Not execution-derived grouping.}  The execution-derived decomposition groups
tasks by neutral summaries of how the initial skill behaved on the training
set.
It is the stronger alternative on SearchQA and LiveMath, yet it still trails
by 6.6, 11.0, and 12.9 points.  Its groups describe how the initial skill
behaved, so they keep pointing at that skill while later rounds change it,
which is consistent with its lower endpoints.

\emph{Not matrix statistics alone.}  The randomized capability assignment
preserves the task and capability degree sequences of $Q$ while breaking which
task carries which requirement.  It reduces accuracy by 11.1, 17.5, and 20.2
points, so which task carries which requirement, and not the degree sequences
alone, drives the gain.

Across the three controls, capabilities read from task specifications give the
strongest final skill on every benchmark, and they are the only source that
stays fixed while the skill changes.

\subsection{Assignment Stability}
\label{app:stability}

We assess stability across three independent proposal calls that read the same
task specifications and do not observe one another's outputs.  After aligning common
behavioral-requirement references, we compare the resulting assignments using
the adjusted Rand index (ARI) and normalized mutual information (NMI).

\begin{table}[!ht]
\centering
\caption{Assignment agreement across three independent proposal calls on the
same task specifications. $K$: proposal-specific capability counts; Common refs.:
aligned behavioral-requirement references; ARI and NMI: adjusted Rand index
and normalized mutual information between proposal assignments.}
\label{tab:stability-appendix}
\small
\setlength{\tabcolsep}{3pt}
\begin{tabular}{@{}lccrrr@{}}
\toprule
Benchmark & Views & $K$ & Common refs. & ARI & NMI \\
\midrule
SearchQA & 3 & 18/13/17 & 47.7 & .883 & .953 \\
SSB & 3 & 17/16/19 & 35.7 & .913 & .976 \\
LiveMath & 3 & 11/11/10 & 33.3 & .973 & .989 \\
IFBench & 3 & 30/24/42 & 112.3 & .912 & .974 \\
\bottomrule
\end{tabular}
\end{table}

The three views propose different numbers of capabilities, yet they assign
tasks to them in much the same way on all four benchmarks (ARI .883--.973; NMI
.953--.989; Table~\ref{tab:stability-appendix}), so the assignments survive a
change of proposal granularity and rest on no single view.

\subsection{Capability-Guided Revision: Full Crossover Analysis}
\label{app:coupling-analysis}

\begin{table}[!ht]
\centering
\caption{Per-benchmark focus--evidence crossover (five historical revision states
per benchmark). Accept: candidate acceptance rate; $\Delta_{\rm sel}$, $\Delta_{\rm tr}$:
paired candidate gains (points) on the selection and training sets.}
\label{tab:crossover-per-benchmark}
\small
\setlength{\tabcolsep}{3pt}
\begin{tabular}{@{}lrrr@{}}
\toprule
Condition & Accept (\%) & $\Delta_{\rm sel}$ & $\Delta_{\rm tr}$ \\
\midrule
\multicolumn{4}{@{}l}{\textit{SearchQA}} \\
TM & 40.0 & $+5.30$ & $+10.50$ \\
T$\neg$M & 20.0 & $+5.30$ & $+3.42$ \\
RM & 40.0 & $+7.50$ & $+3.92$ \\
R$\neg$M & 40.0 & $+3.13$ & $+1.92$ \\
\addlinespace
\multicolumn{4}{@{}l}{\textit{SpreadsheetBench}} \\
TM & 40.0 & $+7.00$ & $+13.33$ \\
T$\neg$M & 20.0 & $+4.17$ & $+7.71$ \\
RM & 40.0 & $+6.00$ & $+8.10$ \\
R$\neg$M & 0.0 & $+5.00$ & $+4.78$ \\
\addlinespace
\multicolumn{4}{@{}l}{\textit{LiveMath}} \\
TM & 40.0 & $0.00$ & $+4.29$ \\
T$\neg$M & 40.0 & $+1.39$ & $+3.77$ \\
RM & 0.0 & $-6.67$ & $+3.98$ \\
R$\neg$M & 20.0 & $-1.85$ & $+3.47$ \\
\bottomrule
\end{tabular}
\end{table}

Choosing the capability to revise and the evidence to read from the same
capability produces the strongest candidates.  The crossover shows this
through two controlled contrasts: matched against mismatched evidence under a
fixed focus policy (TM$-$T$\neg$M and RM$-$R$\neg$M), and top-priority against
randomized focus under matched evidence (TM$-$RM).

\emph{Evidence alignment.}  On the training set, matched evidence raises
candidate gain under top-priority focus by 7.08, 5.62, and 0.52 points on
SearchQA, SpreadsheetBench, and LiveMath, and under randomized focus by 2.00,
3.32, and 0.51 points (Table~\ref{tab:crossover-per-benchmark}).  The
direction holds under both focus policies and on all three benchmarks, so
evidence drawn from the capability under revision beats evidence drawn from
another one whichever capability the round selects.

\emph{Capability prioritization.}  With matched evidence, top-priority focus
exceeds randomized focus by 6.58, 5.23, and 0.31 points, so targeting the
capability with the largest unresolved task mass adds to what matching the
evidence already gives.

\emph{Joint reading.}  TM, the only condition that prioritizes a capability and
reads its own evidence, attains the largest training set candidate gain on
every benchmark and a 40.0\% acceptance rate on each.  Margins are smallest on
LiveMath, where selection set contrasts are also less uniform; the training
set candidate gain serves as the primary crossover comparison
(Appendix~\ref{app:focus-evidence-crossover}).

\begin{table}[!ht]
\centering
\caption{Ablations of capability-guided revision on SearchQA,
SpreadsheetBench, and LiveMath (test accuracy, \%).}
\label{tab:ablation-results}
\small
\setlength{\tabcolsep}{3pt}
\begin{tabular}{@{}lrrr@{}}
\toprule
Variant & SearchQA & SSB & LiveMath \\
\midrule
\textbf{SkillFocus} & \textbf{80.1} & \textbf{77.1} & \textbf{43.6} \\
w/o matched evidence & 71.4 & 58.6 & 20.8 \\
w/o capability prioritization & 68.7 & 38.3 & 20.8 \\
\bottomrule
\end{tabular}
\end{table}

\paragraph{Selected Capabilities in the formal runs.}
Table~\ref{tab:selected-capabilities} records which capabilities the priority
rule of Equation~\ref{eq:profile-priority} actually selected over the five
rounds of each formal run.  The selected capability is not always the broadest
one: SpreadsheetBench repeatedly selects a requirement covering 74\% of its
task population, LiveMath alternates between full-support and partial-support
requirements, and IFBench selects requirements covering between 3\% and 20\%
of its tasks.  Selecting a capability narrows the pool of task records only
when that capability has partial support.  At full support the evidence
population stays the same, and the selected capability still states the
requirement that frames diagnosis and limits the injected revision history to
earlier attempts on that requirement.

\begin{table}[!ht]
\centering
\caption{Capabilities selected by the priority rule over the five rounds of
each formal run. Support is the number of tasks requiring the capability in
the frozen $Q$; Share is its fraction of the task population. A capability is
selected in consecutive rounds when it is the only eligible one.}
\label{tab:selected-capabilities}
\small
\setlength{\tabcolsep}{3pt}
\begin{tabular}{@{}llrrr@{}}
\toprule
Benchmark & capability & Rounds & Support & Share \\
\midrule
SearchQA & $c_1$ & 3 & 600 & 100.0\% \\
SearchQA & $c_2$ & 1 & 600 & 100.0\% \\
SearchQA & $c_3$ & 1 & 289 & 48.2\% \\
\addlinespace
SpreadsheetBench & $c_1$ & 5 & 89 & 74.2\% \\
\addlinespace
LiveMath & $c_1$ & 2 & 53 & 100.0\% \\
LiveMath & $c_3$ & 2 & 31 & 58.5\% \\
LiveMath & $c_7$ & 1 & 24 & 45.3\% \\
\addlinespace
IFBench & $c_{29}$ & 3 & 3 & 3.4\% \\
IFBench & $c_8$ & 1 & 18 & 20.2\% \\
IFBench & $c_{12}$ & 1 & 4 & 4.5\% \\
\bottomrule
\end{tabular}
\end{table}

\paragraph{End-to-end ablations.}
Table~\ref{tab:ablation-results} reports the end-to-end ablations defined in
Appendix~\ref{app:ablation-details}.  Under global evidence, keeping capability prioritization raises the endpoints
by 2.7, 20.3, and 0.0 points on SearchQA, SpreadsheetBench, and LiveMath;
with prioritization kept, matching the evidence to the selected capability
adds 8.7, 18.5, and 22.8 points.  LiveMath separates the two most sharply:
prioritization under global evidence recovers none of the loss, while matched
evidence recovers 22.8 points.  These ablations rerun the whole evolution
trajectory, so they measure what each component contributes to the final
skill, alongside the same-state crossover above.

\subsection{Behavioral Scope of Capability-Guided Revisions}
\label{app:analysis-localization}

A whole-skill edit can change tasks beyond the ones that selected its
capability.  Table~\ref{tab:edit-response} measures how far each recorded
change concentrates on target tasks.

\begin{table}[!ht]
\centering
\caption{Target and non-target edit responses. T/N: group sizes;
$\Delta_T/\Delta_N$: net changes (points); $R_d(e)$: their difference;
Screen./Acc.: screening record/accepted candidate (Acc.-1: first accepted).}
\label{tab:edit-response}
\small
\setlength{\tabcolsep}{3pt}
\begin{tabular}{@{}llccr@{}}
\toprule
Benchmark & Record & T/N $n$ & $\Delta_T/\Delta_N$ & $R_d(e)$ \\
\midrule
SQA & Screen. & 93/107 & $+1.1/+0.0$ & $+1.1$ \\
SSB & Acc.-1 & 89/31 & $+46.1/+16.1$ & $+29.9$ \\
LM & Acc. & 31/22 & $+6.5/+4.5$ & $+1.9$ \\
IFB & Acc. & 3/86 & $+0.0/+8.1$ & $-8.1$ \\
\bottomrule
\end{tabular}
\end{table}

The target behaviors in the recorded responses are selecting the answer type
requested by a clue, writing the exact formula in specified answer cells,
selecting the strongest claim supported by a prompt, and ending each required
sentence with an emoji. Target and non-target groups are defined by the frozen
task--capability Matrix. $R_d(e)$ is the target-minus-non-target difference in
net percentage-point change.

SpreadsheetBench shows the clearest targeted response
(Table~\ref{tab:edit-response}), consistent with a specific target-cell
contract. SearchQA and LiveMath show smaller positive
contrasts ($+1.1$ and $+1.9$ percentage points), indicating that reusable skill
rules can benefit both target and neighboring behaviors. IFBench shows a negative contrast ($-8.1$ percentage points), with its improvement
appearing mainly outside a three-task target group.

Appendix~\ref{app:representative-revision} examines the first accepted
SpreadsheetBench edit in detail. The second accepted edit has a positive target
response together with a larger non-target response, recording the collateral
changes possible when whole-skill instructions affect overlapping behaviors.
The IFBench response contains three target tasks and a much larger non-target group;
its support size is reported with the response value.

A selected capability therefore decides which evidence a round reads and which
direction the edit takes; it does not fence in where the edit lands.  Whole-skill
screening is what keeps the aggregate change positive.

\subsection{Selection Screening and Held-out Transfer}
\label{app:selection-transfer}

The selection set screens candidates inside the loop, while test accuracy
reports what the retained skill transfers to unseen tasks.
Table~\ref{tab:selection-test} lists both endpoints for SkillFocus and
SkillOpt.  SkillFocus builds its selection set for capability coverage and
remaining headroom, so the set collects tasks the initial skill leaves
unresolved, and its accuracy stays below the corresponding test accuracy on
three of the four benchmarks: 49.5 against 80.1 on SearchQA, 38.9 against 43.6
on LiveMath, and 77.8 against 82.0 on IFBench.  A set built this way is meant
to score low.

\begin{table}[!ht]
\centering
\caption{Final selection set and held-out test accuracy (\%). The two columns
cover different task sets, so their difference is not a calibrated
generalization gap across methods.}
\label{tab:selection-test}
\small
\setlength{\tabcolsep}{4pt}
\begin{tabular}{@{}lrrrr@{}}
\toprule
& \multicolumn{2}{c}{SkillFocus} & \multicolumn{2}{c}{SkillOpt} \\
\cmidrule(lr){2-3}\cmidrule(lr){4-5}
Benchmark & Sel. & Test & Sel. & Test \\
\midrule
SearchQA & 49.5 & 80.1 & 76.0 & 76.1 \\
SpreadsheetBench & 80.0 & 77.1 & 85.0 & 62.1 \\
LiveMath & 38.9 & 43.6 & 44.4 & 41.1 \\
IFBench & 77.8 & 82.0 & 89.7 & 73.9 \\
\bottomrule
\end{tabular}
\end{table}

SkillOpt shows the opposite ordering on SpreadsheetBench and IFBench, where
its screening accuracy of 85.0 and 89.7 exceeds its test accuracy of 62.1 and
73.9 by 22.9 and 15.8 points.  SkillFocus screens lower on both benchmarks and still reaches the higher test
accuracy.  A screening score reports how well a candidate fits the tasks it
was retained on, and it carries no guarantee about unseen tasks.

The two columns cover different task sets, and the selection sets differ in
composition across methods, so the screening-to-test difference carries no
cross-method claim about generalization.  What the table does support is how
to read the trajectories in Figure~\ref{fig:evolution-trajectories}, which
plot selection set gain along the search path, and the construction control
below, which measures what a different selection set retains.

\subsection{Evolution Cost and Selection-Set Design}
\label{app:analysis-efficiency}

\begin{figure*}[t]
\centering
\includegraphics[width=0.96\textwidth]{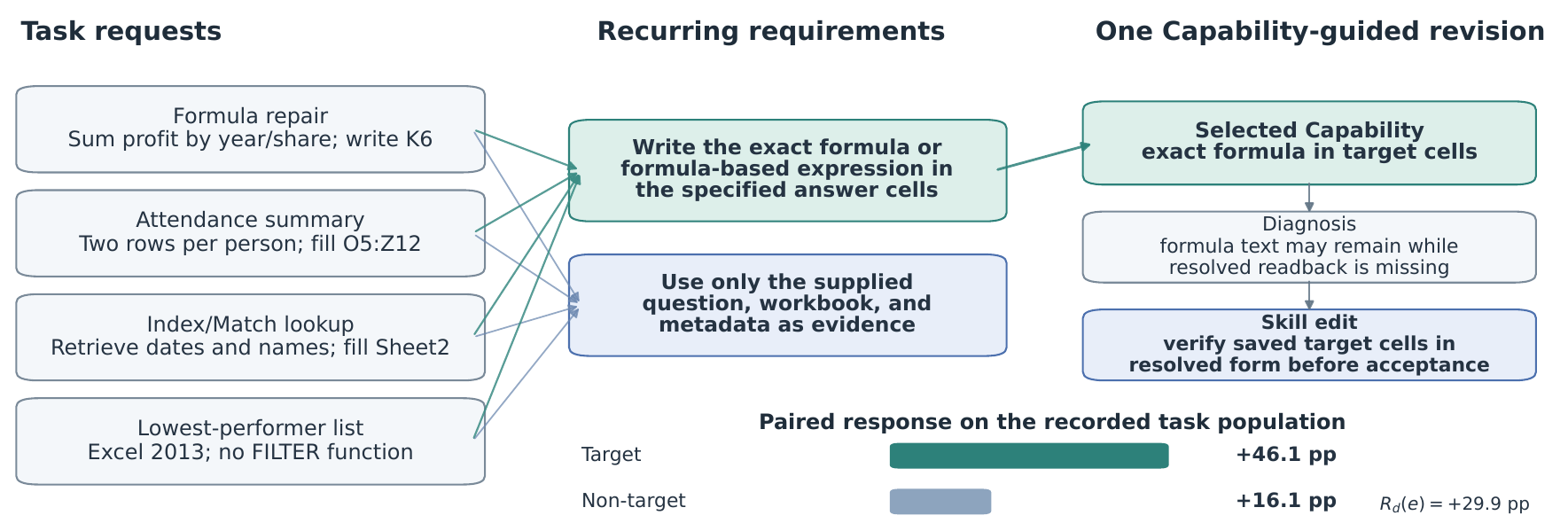}
\caption{Representative capability-guided revision on SpreadsheetBench. The
case connects diverse task requests to a recurring behavioral requirement, an
evidence-grounded revision, and paired target/non-target responses.}
\label{fig:qualitative-capability-case}
\end{figure*}

\begin{figure*}[t]
\centering
\begin{minipage}{0.97\textwidth}
\centering
\begin{Verbatim}[fontsize=\small,frame=single,framesep=4pt]
4. Confirm the target cells/range contain the expected values.
\end{Verbatim}
\smallskip
\begin{Verbatim}[fontsize=\small,frame=single,framesep=4pt]
Target contract: verify the requested result in the task-specified target cells or ranges.

4. Confirm the target cells/range by reopening or rereading the saved
output and checking the concrete saved values, not only the formula string.
\end{Verbatim}
\end{minipage}
\caption{Representative accepted skill revision on SpreadsheetBench.
The first excerpt is the current instruction; the second is the accepted
candidate's added target-cell verification and read-back contract.}
\label{fig:skill-revision-excerpt}
\end{figure*}

SkillFocus spends its evolution budget on diagnosing one requirement at a time
and on rejecting weak candidates early, which
Figure~\ref{fig:evolution-trajectories} and the token ledger in
Table~\ref{tab:token-accounting} show from two sides.  The curves plot
selection set accuracy gain against cumulative evolution tokens instead of
round counts: horizontal segments are rejected candidates that consume tokens
without changing the retained skill, and jumps follow accepted updates.  The
ledger separates one-time initialization, iterative evolution, and final test
set inference.

The evolution cost can be read as
\[
\begin{aligned}
C_{\mathrm{evo}}={}&C_{\mathrm{base}}+C_{\mathrm{induction}}\\
&+C_{\mathrm{revision}}+C_{\mathrm{validation}},
\end{aligned}
\]
where $C_{\mathrm{base}}$ is the initial evaluation of the current skill,
$C_{\mathrm{induction}}$ is the one-time cost of inducing and
freezing $(\mathcal C,Q)$, $C_{\mathrm{revision}}$ covers per-round diagnosis
and edit generation, and $C_{\mathrm{validation}}$ covers candidate screening, training set
re-evaluation, and recomputation of the capability scores.  In
Table~\ref{tab:token-accounting},
Initialization aggregates $C_{\mathrm{base}}+C_{\mathrm{induction}}$,
while Iterative Evolution aggregates
$C_{\mathrm{revision}}+C_{\mathrm{validation}}$.  SkillFocus pays $C_{\mathrm{induction}}$ once.
In exchange, every round reads the capability scores, the selected capability,
and its evidence off $Q$ and the current execution record, and never re-derives
which tasks and failures belong together.  Screening on the selection set first
further limits $C_{\mathrm{validation}}$: a candidate that fails there is
rejected without training set re-evaluation, and an accepted candidate's
training set record becomes the next execution record.  The saving comes from
where tokens are spent, not from making fewer model calls.

Appendix~\ref{app:selection-transfer} shows what the constructed selection set
screens on; the control below shows what a different one retains.  Replacing
the constructed selection set with uniform random sampling of the same size
(Appendix~\ref{app:selection-control}) lowers the final test accuracy on
all three evaluated benchmarks, by 5.1 points on SearchQA, 3.9 points on
SpreadsheetBench, and 17.2 points on LiveMath.  Which tasks screen the
candidates decides which skill survives the loop: tasks chosen for capability
coverage, feature diversity, and diagnosable failure retain different
candidates than a uniform sample, and every retained candidate changes the
skill, the next execution record, and the next capability priority.  A
selection set built for coverage and headroom keeps more of the updates that
transfer beyond the training evidence, and the effect is largest on LiveMath,
where the set is smallest.  A selection set constructed independently of the
priority rule would separate its composition from capability prioritization
directly; the uniform random sample is the closest available approximation,
and a dedicated control remains open.

\subsection{Representative Capability-Guided Revision}
\label{app:representative-revision}

We present one complete revision chain from the recorded SpreadsheetBench run,
following the same objects SkillFocus works with: a recurring requirement, the
evidence matched to it, a diagnosis, a whole-skill revision, and paired
candidate responses.

\paragraph{Cross-task Capability and evidence.}
The four requests in Figure~\ref{fig:qualitative-capability-case} have
different surface goals, including formula repair, attendance summarization,
lookup construction, and a version-constrained list. They share one requirement: write the exact formula or formula-based
expression into the specified answer cells. The frozen $Q$ assigns that
requirement to the same capability across the displayed rows, so one revision
can be evaluated on tasks that look unrelated.

\paragraph{Evidence-grounded diagnosis.}
\noindent\fbox{\begin{minipage}{0.96\columnwidth}
\small
\textbf{Observed pattern.} Executions under the current skill may write a
formula string without verifying the resolved value in the requested target
cells.

\medskip
\textbf{Diagnosis.} The skill lacks an explicit post-save check that reopens the
output and verifies the concrete values in the task-specified cells.

\medskip
\textbf{Editable location.} Extend the target-cell confirmation instruction
with an identity check and a read-back check before the task output is
considered complete.\end{minipage}}

The evidence bundle combines failures on the target capability, a successful
contrast, related current-skill content, and earlier revisions of the same
capability. The diagnosis ties this evidence to an observed execution pattern
and to one place in the skill. The resulting edit is applied to the complete
skill before candidate screening.

\paragraph{Skill edit.}
The accepted edit adds a target contract and strengthens the confirmation step
of the targeted instruction (Figure~\ref{fig:skill-revision-excerpt}); the
complete skill files are omitted for brevity.

\paragraph{Paired candidate response.}

\begin{table}[!ht]
\centering
\caption{Paired response of the representative SpreadsheetBench revision.
Repair/reg.\ counts failure-to-success and success-to-failure changes; the
target-minus-non-target contrast is $+29.9$ points.}
\label{tab:prompt-skill-response}
\small
\setlength{\tabcolsep}{3pt}
\begin{tabular}{@{}lrrr@{}}
\toprule
Group & $n$ & Repair/reg. & Net $\Delta$ \\
\midrule
Target & 89 & 42/1 & $+46.1\%$ \\
Non-target & 31 & 7/2 & $+16.1\%$ \\
\bottomrule
\end{tabular}
\end{table}

The case materials thus run from capability-matched evidence to a diagnosis, a
whole-skill edit, and a paired candidate evaluation.

\paragraph{Why the Capability matters in this case.}
The four tasks differ in surface form, yet their failures expose the same
target-cell contract. A task-local repair could specialize to one spreadsheet
operation, whereas capability-matched evidence pools failures and successes
across these tasks before diagnosis. The revision therefore changes a reusable
verification rule instead of adding an instance-specific workaround, and it
moves target tasks more than non-target ones
(Table~\ref{tab:prompt-skill-response}).

\section{Experimental Details}
\label{app:evaluation-details}

This section documents how the experiments are constructed; the corresponding
observations are reported in Section~\ref{sec:experiments} and
Appendix~\ref{app:additional-analysis}.

\subsection{Benchmarks and Data Split}
\label{app:benchmarks}
\label{app:split}

\begin{table}[!ht]
\centering
\caption{Benchmarks used in the experiments. Task format states what the agent
reads and produces; Evaluation states the deterministic checker applied to
each response. }
\label{tab:benchmark-details}
\small
\setlength{\tabcolsep}{3pt}
\begin{tabularx}{\columnwidth}{@{}l>{\raggedright\arraybackslash}X>{\raggedright\arraybackslash}p{0.27\columnwidth}@{}}
\toprule
Benchmark & Task format & Evaluation \\
\midrule
SearchQA & Jeopardy!-style clues answered in a closed-context, single-turn setting & Deterministic answer matching \\
SpreadsheetBench & Spreadsheet manipulation on the Verified subset; the agent reads and writes workbook files & Target cells compared with the reference workbook \\
LiveMath & Multiple-choice mathematical reasoning with deterministically shuffled options; code execution is disabled & Exact match of the selected option \\
IFBench & Single-turn free-text responses under verifiable output constraints & Programmatic constraint verification \\
\bottomrule
\end{tabularx}
\end{table}

Table~\ref{tab:benchmark-details} summarizes the four benchmarks.  SearchQA
tests whether the agent returns the answer type and entity requested by a
clue.  SpreadsheetBench requires writing correct formulas or values into the
specified cells of a workbook.  LiveMathematicianBench requires selecting the
strongest mathematical claim among shuffled options without code execution.
IFBench constrains responses along count, format, ratio, repetition,
sentence, word, and custom dimensions.  Each benchmark evolves an independent
skill from an initial skill that mirrors the SkillOpt setting for that
benchmark, all methods share the same
evaluator, and every reported score is accuracy on the held-out test set.

The formal protocol uses
\[
|D_{\rm tr}|:|D_{\rm sel}|:|D_{\rm test}|=2:1:7.
\]
Let $\mathcal D$ denote the task pool after removing the held-out test set.
SkillFocus induces and freezes its capability space over $\mathcal D$, executes
the initial skill on the same pool to obtain $O_0$, and then constructs
$(D_{\rm sel},D_{\rm tr})$ from $(\mathcal C,Q,O_0)$ as specified in
Section~\ref{sec:candidate-selection}.  The construction prioritizes
capability coverage, task diversity, and diagnosable improvement opportunity;
a reserve of tasks solved by $S_0$ keeps regressions visible.  Both sets are
fixed before the first round.  selection outcomes enter the loop through
candidate screening, and the test set stays closed until final evaluation.

\subsection{Baselines and Fairness}
\label{app:baselines}

\emph{No Skill} runs the target model on the benchmark task without
procedural guidance.  \emph{One-shot Skill} applies a single skill document
created once before evaluation, without iterative execution feedback.
\emph{Trace2Skill} \citep{ni2026trace2skill} executes the target agent on the
training set, lets independent optimizer analysts propose local skill patches
from individual trajectories, and consolidates the patches into one skill
document with a final optimizer call.  \emph{GEPA} \citep{agrawal2026gepa}
treats the skill text as a single optimizable component and applies
reflective mutation with Pareto-based candidate selection, a reflection
minibatch of three examples, and a budget of 160 metric calls.
\emph{SkillOpt} \citep{yang2026skillopt} uses its official implementation to
iteratively optimize a retained skill.  The skill-optimization baselines use
the same evaluators, target model, optimizer model, and initial skill as
SkillFocus.  Candidate retention remains method-specific: GEPA keeps its
Pareto-based selection, SkillOpt its retained-skill update, and SkillFocus
its paired selection and training check.  Fairness is enforced through the same task pool and
held-out test split, evaluators, target and optimizer models, and initial
skills where applicable.  Optimization budgets follow each method's native
protocol, while how each method splits the pool and retains candidates remains
part of its design.  Table~\ref{tab:baseline-fairness} summarizes these
shared and method-specific conditions.

\begin{table*}[t]
\centering
\caption{Shared and method-specific conditions across compared methods. ``--''
denotes a condition that does not apply to the method.}
\label{tab:baseline-fairness}
\small
\setlength{\tabcolsep}{3pt}
\begin{tabularx}{\textwidth}{@{}>{\raggedright\arraybackslash}p{0.13\textwidth}*{6}{>{\raggedright\arraybackslash}X}@{}}
\toprule
Condition & No Skill & One-shot Skill & Trace2Skill & GEPA & SkillOpt & SkillFocus \\
\midrule
Evaluator and test split & Shared & Shared & Shared & Shared & Shared & Shared \\
Target model & DeepSeek-V4-Flash & DeepSeek-V4-Flash & DeepSeek-V4-Flash & DeepSeek-V4-Flash & DeepSeek-V4-Flash & DeepSeek-V4-Flash \\
Initial skill & -- & -- & Shared & Shared & Shared & Shared \\
Optimizer & -- & -- & GPT-5.4 & GPT-5.4 & GPT-5.4 & GPT-5.4 \\
Evolution budget & -- & -- & One analysis and consolidation pass & 160 metric calls & Official setting & At most five rounds, one screened candidate per round \\
Candidate retention & -- & -- & Single consolidated skill & Pareto-based selection & Retained-skill update & Positive paired change on selection and training sets \\
Evolution cost reported & -- & -- & -- & -- & Yes & Yes \\
\bottomrule
\end{tabularx}
\end{table*}

\subsection{Models and Inference Settings}
\label{app:configuration}

The formal runs use GPT-5.4 as the optimizer model and
DeepSeek-V4-Flash as the fixed target model, with random seed 42 and
64 within-run workers. The target model runs with temperature 0 and
thinking enabled at low reasoning effort, and optimizer calls use temperature 0. The evolution horizon is at most five rounds with one
candidate per round. The same test set is
used for final comparison and remains unseen during evolution.

\subsection{Evolution Cost Accounting}
\label{app:token-accounting}

\begin{table}[!ht]
\centering
\caption{SkillFocus inference-token ledger (millions). SQA, SSB, LM, and IFB
abbreviate the four benchmarks; C/A: proposed/accepted candidates; Init.,
Iter.: initialization and iterative evolution.}
\label{tab:token-accounting}
\small
\setlength{\tabcolsep}{3pt}
\begin{tabular}{@{}lrrrrr@{}}
\toprule
Benchmark & C/A & Init. & Iter. & $C_{\mathrm{evo}}$ & Test \\
\midrule
SQA & 5/2 & 5.10M & 15.55M & 20.66M & 8.51M \\
SSB & 5/2 & 6.45M & 22.61M & 29.06M & 4.73M \\
LM & 5/2 & 0.38M & 5.98M & 6.36M & 3.40M \\
IFB & 5/1 & 0.97M & 2.48M & 3.45M & 1.58M \\
\bottomrule
\end{tabular}
\end{table}

\begin{table}[!ht]
\centering
\caption{Pre-final-test evolution cost (millions of inference tokens) of
SkillFocus and SkillOpt. The final row averages the per-benchmark relative
changes.}
\label{tab:appendix-cost-comparison}
\small
\setlength{\tabcolsep}{3pt}
\begin{tabular}{@{}lrrr@{}}
\toprule
Benchmark & SkillFocus & SkillOpt & Relative change \\
\midrule
SearchQA & 20.66 & 45.21 & $-54.3\%$ \\
SpreadsheetBench & 29.06 & 40.55 & $-28.3\%$ \\
LiveMath & 6.36 & 8.14 & $-21.9\%$ \\
IFBench & 3.45 & 3.21 & $+7.5\%$ \\
\midrule
Mean & -- & -- & $-24.0\%$ \\
\bottomrule
\end{tabular}
\end{table}

SkillFocus costs in Table~\ref{tab:appendix-cost-comparison} are the sum of
Initialization and Iterative Evolution in Table~\ref{tab:token-accounting}.
SkillOpt costs are reconstructed by summing prompt and completion tokens over
every recorded optimization step in the matched run history, with final test
set inference excluded.  The relative change is
$(C_{\mathrm{evo}}^{\mathrm{SkillFocus}}-C_{\mathrm{evo}}^{\mathrm{SkillOpt}})
/C_{\mathrm{evo}}^{\mathrm{SkillOpt}}$, so negative values indicate a
reduction; the mean over benchmarks gives the 24\% reduction reported in the
main text.  Initialization is the one-time pre-evolution bucket: the baseline evaluation
and the capability induction that constructs and freezes $(\mathcal C,Q)$.
Iterative evolution contains per-round diagnosis, edit generation, screening,
recomputation of the capability scores, and memo calls.  Both buckets constitute
$C_{\mathrm{evo}}$; final test set inference is reported separately as
$C_{\mathrm{test}}$ and is excluded from $C_{\mathrm{evo}}$.  Evolution tokens
include model calls incurred before the final skill is fixed, including
rejected and abstained candidates; a round that attempts more than one
capability counts every attempt.  Values are rounded to millions of inference tokens.

\subsection{Analysis Protocols}
\label{app:analysis-protocols}

The following protocols specify how the ablations and analyses are constructed;
their results appear in Section~\ref{sec:experiments} and
Appendix~\ref{app:additional-analysis}.

\subsubsection{Ablation Controls}
\label{app:ablation-details}

\emph{w/o capability prioritization} drops the priority rule and supplies
global training set evidence. \emph{w/o matched evidence} keeps the
priority rule and supplies global training set evidence.
\emph{randomized capability assignment} preserves the task and capability
degree sequences while randomizing their correspondence. The three
interventions separate where the capabilities come from, which one a round
selects, and which evidence the revision reads.

\subsubsection{Counterfactual Focus--Evidence Crossover}
\label{app:focus-evidence-crossover}

The focus--evidence crossover is a post-hoc mechanism analysis over the same
historical revision states; it does not change capability induction, the
evolution horizon, candidate gating, or held-out evaluation. Each unit is a
candidate-producing state
\[
X_t=(S_t,O_t,\mathbf q^t,\mathcal H_t;\mathcal C,Q).
\]
The TM condition reuses the historical top-priority focus and its matched
evidence. T$\neg$M keeps the top-priority focus but replaces its evidence with
the evidence for a code-selected non-top capability $c_r$. RM uses $c_r$ for
both focus and evidence, while R$\neg$M uses $c_r$ for focus and the historical
top capability's evidence. The same $c_r$ is used for all three counterfactual
branches of a state and is selected without reading their outcomes.

All four conditions share the current skill state, frozen $(\mathcal C,Q)$,
state-bound task sets, model and prompt configuration, evidence budget,
candidate generation, and screening protocol. Counterfactual branches do not
write back to the historical run or to one another. The reported acceptance rate,
$\Delta_{\rm sel}$, and $\Delta_{\rm tr}$ are computed against the same
historical current skill, and Table~\ref{tab:focus-evidence-crossover} reports
macro averages over five states per benchmark. Selection set outcomes remain a screening
measure, so the training set candidate gain is the clearest comparison.  The crossover measures
candidate-level changes from historical states, not held-out test
accuracy, and it compares controlled contrasts instead of estimating a formal
factorial interaction effect.  Per-benchmark results are reported in
Appendix~\ref{app:coupling-analysis}.

\subsubsection{Decomposition-Source Control}
\label{app:decomposition-source}

To isolate where the capabilities come from, we hold the downstream evolution
interface fixed.  The SkillFocus condition uses the formal capability space
and its endpoint.  The task-semantic condition reads the same pre-execution
task specifications and describes each task by its overall objective,
operation, and domain.  The execution-derived condition summarizes how the
initial skill behaved on the full non-test task pool.

Let $\mathcal D$ denote the task pool after the test set has been removed.  The
SkillFocus and task-semantic arms induce and freeze $(\mathcal C,Q)$ from
$\mathcal D$ before the initial skill is executed.  The initial skill is then executed on
the complete pool,
\[
O_0=\operatorname{Execute}(A,S_0,\mathcal D),
\]
and $(\mathcal C,Q)$ together with $O_0$ are passed to the shared selection
constructor,
\[
(D_{\rm sel},D_{\rm tr})
=\operatorname{ConstructSelection}(\mathcal C,Q,O_0,M).
\]
For the execution-derived condition, this full-pool $O_0$ supplies the input to
the behavior summary, which is frozen as $(\mathcal C,Q)$ before the same
constructor assigns the two roles.  The subsequent evolution uses $O_0$
restricted to $D_{\rm tr}$.  Every arm forms $(\mathcal C,Q)$ before the split,
so the training and selection sets follow from the frozen $(\mathcal C,Q)$ and
$O_0$.

Each arm freezes its $(\mathcal C,Q)$ once before round 0 and keeps it for the
full run.  The alternatives are allowed the same number of capabilities, with
$K^{\mathrm{sem}}=K^{\mathrm{exec}}=K^{\mathrm{cap}}$, assign several
capabilities per task, apply the same support threshold $n_{\min}=3$, and pass
the same natural-language descriptions to revision, assignment and filtering,
$J_D$, and $J_G$.  They share the same task pool and
held-out test split, initial skill, target and optimizer models, evaluator,
inference settings, evidence budget, prompts, edit operations, history
mechanism, five-round horizon, candidate screening, training re-evaluation,
and paired acceptance rule.  Each condition constructs its own
$(D_{\rm sel},D_{\rm tr})$ from its own $(\mathcal C,Q)$ and $O_0$, so the
selection set matches the capability source under test.  The primary outcome is final held-out test accuracy, reported in
Table~\ref{tab:representation-endpoints}; token cost and candidate counts are
outside this control.

\subsubsection{Selection-Set Construction Control}
\label{app:selection-control}

The control keeps the selection set size and replaces the construction of
Section~\ref{sec:candidate-selection} with uniform random sampling from the
same pool on SearchQA, SpreadsheetBench, and LiveMath.  The rest of the
pipeline is unchanged.  Results are reported in
Appendix~\ref{app:analysis-efficiency}.

\section{Method Details}
\label{app:method-details}
\label{app:pseudocode}

This section details each stage of the main-text algorithm, using the notation
of Section~\ref{sec:method};
$\Delta_X(S',S)$ is the paired change of Equation~\ref{eq:paired-gain}.  The
prompt interfaces behind the model calls are described in
Appendix~\ref{app:prompt-templates}.

\setcounter{algorithm}{0}
\renewcommand{\thealgorithm}{\thesection.\arabic{algorithm}}

\subsection{Capability Induction}
\label{app:algorithm-induction}

Capability induction reads task records from the training and selection sets.
Each record contains the task instruction, input structure, and environment
contract. The extraction interface never sees the current skill, an execution
trace, a success label, a failure explanation, or a candidate edit. The test set remains unavailable until the
final retained skill is fixed.

\begin{algorithm}[!ht]
\centering
\caption{Capability Induction}
\label{alg:app-induction}
\small
\begin{algorithmic}[1]
\Require Task records for the task pool $\mathcal D$
\Ensure Frozen capability set $\mathcal C$ and matrix $Q$
\ForAll{$\tau_i\in\mathcal D$}
    \State $R_i\gets$ requirements of task $i$
\EndFor
\For{$j\gets1$ to $3$}
    \State $P_j\gets$ proposal from $\{R_i\}$ under $\rho_j$
\EndFor
\State $\mathcal C_0\gets$ consensus of $P_1,P_2,P_3$
\State $Q_0\gets$ assignment of each $R_i$ to $\mathcal C_0$
\ForAll{$c_d\in\mathcal C_0$}
    \If{$\sum_{\tau_i\in\mathcal D}Q_{0,i,d}<n_{\min}$}
        \State Remove $c_d$ and unassign its requirements
    \EndIf
\EndFor
\State $(\mathcal C,Q)\gets$ freeze $(\mathcal C_0,Q_0)$
\State \Return $(\mathcal C,Q)$
\end{algorithmic}
\end{algorithm}

Algorithm~\ref{alg:app-induction} extracts requirements from the task records,
forms three independent proposal views, consolidates their registries, and
assigns each requirement to one retained capability or
\textsc{unassigned}. Support filtering runs deterministically before
$(\mathcal C,Q)$ is frozen. The three calls share the same canonical required-
behavior input and use fixed Requirement-, Intervention-, and Boundary-centric
perspectives, respectively; they do not see one another's outputs.
Here $\rho_j$ denotes the perspective injected for proposal call $j$.

\subsection{Selection-Set Construction}
\label{app:selection-construction}

\begin{algorithm}[!ht]
\caption{ConstructSelection}
\label{alg:construct-selection}
\small
\begin{algorithmic}[1]
\Require Task pool $\mathcal D$, frozen $(\mathcal C,Q)$, initial record $O_0$,
         selection size $M$
\Ensure selection set $D_{\rm sel}$ and training set $D_{\rm tr}$
\State $\pi_d\gets$ support share of $c_d$ in $\mathcal D$;\ \ $o_i\gets$ diagnosable failure rate of $i$
       in $O_0$;\ \ $R\gets\{i:O_0$ succeeds on all runs of $i\}$
\State $D_{\rm sel}\gets\emptyset$; $\mathcal K\gets\emptyset$
\While{$|D_{\rm sel}|<M$}
    \State $A\gets\{i\in\mathcal D\setminus D_{\rm sel}$ that leaves every $c_d$
           supported outside $D_{\rm sel}\}$, restricted to $R$ once the
           remaining slots are needed for the reserve
    \State $\gamma_i\gets\sum_{d:\,Q_{i,d}=1,\,c_d\notin\mathcal K}\pi_d$;\ \
           $\nu_i\gets|\text{new features of }i|$
    \State $i^\ast\gets\arg\max_{i\in A}(\gamma_i,\nu_i,o_i)$
           \Comment{lexicographic}
    \State $D_{\rm sel}\mathrel{+}=\{i^\ast\}$;\ \
           $\mathcal K\mathrel{+}=\{c_d:Q_{i^\ast\!,d}=1\}$
\EndWhile
\State \Return frozen sets $(D_{\rm sel},\,\mathcal D\setminus D_{\rm sel})$
\end{algorithmic}
\end{algorithm}

Algorithm~\ref{alg:construct-selection} builds the selection set of
Section~\ref{sec:candidate-selection}.  The coverage weight $\pi_d$ is the
share of pool tasks whose row in $Q$ marks $c_d$, so the same support share
that weights unresolved task mass in Equation~\ref{eq:profile-priority} also
weights coverage here.  Diversity counts the static task features an item adds
to the partial selection set, using the task family, input structure, and
environment contract labels recorded with each item.  The
improvement-opportunity term is the fraction of an item's initial executions
that fail for diagnosable task-logic reasons.  That classification is
deterministic and uses no model call: failed executions are grouped first by
the evaluator's error signal and then by the repair signature of the trace,
and failures caused by the execution environment are counted apart from them,
so environment faults win no slots.  Items solved on every initial execution
form the stable-success reserve, which the construction fills to one fifth of
the selection set once the remaining slots would otherwise leave it
unreachable; this fraction is fixed in advance and is not tuned against any
outcome.  The feasibility constraint keeps, for
every capability, at least one supporting task in the training set.  It reads
$Q$ alone, so the task it keeps need not be one the initial skill fails, and a
capability whose failing tasks all move to the selection set leaves the eligible
capability set of Equation~\ref{eq:profile-priority} for the rest of the run.  Ties at every level are broken by task identifier, which
makes the construction deterministic given the pool, $(\mathcal C,Q)$, and the
initial execution record.

\subsection{Capability Scores and Capability Selection}
\label{app:algorithm-focus}

\begin{algorithm}[!ht]
\centering
\caption{Capability Scores and Capability Selection}
\label{alg:app-focus}
\small
\begin{algorithmic}[1]
\Require $A$, skill $S_t$, record $O_t$, $D_{\rm tr}$, $D_{\rm sel}$, $\mathcal C$, $Q$
\Ensure Record $O_t$, capability scores $\mathbf q^t$, ordered eligible capabilities $\mathcal U_t$
\If{$O_t$ is unavailable}
    \State $O_t\gets$ execute $A$ with $S_t$ on $D_{\rm tr}$
\EndIf
\ForAll{$c_d\in\mathcal C$}
    \State $\pi_d\gets\frac{1}{|D_{\rm tr}\cup D_{\rm sel}|}\sum_{i\in D_{\rm tr}\cup D_{\rm sel}}Q_{i,d}$
    \State $q_d^t\gets$ mean of $Y_i(S_t)$ over $i\in D_{\rm tr}$ with $Q_{i,d}=1$
    \State $m_d^t\gets\pi_d\left(1-q_d^t\right)$
\EndFor
\State $\mathcal U_t\gets\{c_d:$ diagnosable failure in $D_{\rm tr}$, support in $D_{\rm sel}\}$
\State \Return $(O_t,\mathbf q^t,\mathcal U_t$ by decreasing $m_d^t)$
\end{algorithmic}
\end{algorithm}

Algorithm~\ref{alg:app-focus} recomputes the capability scores from the
training set and orders the eligible capabilities by unresolved task mass, with
ties broken by registry order; a round tries them in this order until one
yields a candidate.  A capability whose last candidate was rejected for a
confirmed non-positive paired effect sits out the next round, and returns at
once when no other capability is eligible; in the reported runs every round
produced a candidate for its first eligible capability. An accepted candidate's
re-evaluation becomes the next $O_t$.

\subsection{Capability-Guided Revision}
\label{app:algorithm-revision}

For the selected capability, retrieval keeps the training set tasks that $Q$
marks for it. The bundle contains failures on those tasks, successful
executions to compare against, related current-skill content, and earlier edits
made for the same capability. Failure examples are
stratified by observed reason, duplicate traces are removed, and retrieval
order rotates across rounds when more examples are available than the evidence
allowance. A failed task is paired first with a successful execution of the same task
when available, otherwise with a successful task requiring the selected
capability.  When the record holds no successful execution for that capability,
the diagnosis is told so instead of receiving a substitute contrast.

\begin{algorithm}[!ht]
\centering
\caption{Capability-Guided Revision}
\label{alg:app-revision}
\small
\begin{algorithmic}[1]
\Require Selected capability $c_t^\star$, skill $S_t$, record $O_t$, history $\mathcal H$, $Q$
\Ensure Candidate $S'_t$ and record $r_t$, or \textsc{abstain}
\State $I_t\gets\{i\in D_{\rm tr}:Q_{i,d_t^\star}=1\}$
\State $\mathcal E_t\gets$ failures and successes on $I_t$ in $O_t$
\State Add skill content related to $c_t^\star$ to $\mathcal E_t$
\State Add history entries of $c_t^\star$ in $\mathcal H$ to $\mathcal E_t$
\For{$j\gets1$ to $J_D$}
    \State $z_t^{(j)}\gets$ diagnosis from $\mathcal E_t$ under $\pi^D_j$
\EndFor
\State $z_t\gets$ integration of $\{z_t^{(j)}\}$ against $\mathcal E_t$
\If{$z_t$ is not actionable}
    \State \Return $(\textsc{abstain},r_t=(z_t))$
\EndIf
\For{$j\gets1$ to $J_G$}
    \State $g_t^{(j)}\gets$ edit concept for $z_t$ under $\pi^G_j$
\EndFor
\State $e_t\gets$ integration of $\{g_t^{(j)}\}$ into one edit
\If{no valid edit exists}
    \State \Return $(\textsc{abstain},r_t=(z_t))$
\EndIf
\State $S'_t\gets$ $S_t$ with $e_t$ applied; $r_t\gets(z_t,e_t)$
\State \Return $(S'_t,r_t)$
\end{algorithmic}
\end{algorithm}

Algorithm~\ref{alg:app-revision} makes the evidence boundary explicit: the
selected capability picks the failures and success contrasts, pulls in the
related skill content and its own revision history, and then frames diagnosis
and edit integration. The edit is applied to the complete skill.  Before execution, a
candidate is discarded if it cites evidence outside its bundle, targets an
instruction the skill does not contain, duplicates an existing instruction, or
copies task-specific entities such as task identifiers, file or sheet names,
and target coordinates.  A removal is admitted only when observations across
several rounds and tasks associate the targeted instruction with the failing
behavior.  These checks are deterministic and add no model calls.
In the formal runs, $J_D=J_G=3$.  The value follows the three frozen operational perspectives of each stage,
with one independent Explorer per perspective, and was set without reference
to benchmark outcomes; sensitivity to the number of Explorers is left to
future work.  Each Explorer call is independent,
uses its corresponding configured perspective $\pi^D_j$ or $\pi^G_j$, and is
followed by one Integrator call for that stage.  The independent-view design
follows the logic of self-consistency \citep{wang2023selfconsistency}; its
multi-agent consolidation step is also related to debate-based aggregation
\citep{du2024multiagentdebate}.

\subsection{Candidate Screening}
\label{app:algorithm-update}

Candidate screening compares the candidate and the current skill on identical
task instances.

\begin{algorithm}[!ht]
\centering
\caption{Candidate Screening and Skill \mbox{Update}}
\label{alg:app-update}
\small
\begin{algorithmic}[1]
\Require $A$, skill $S_t$, candidate $S'_t$, record $O_t$, $D_{\rm sel}$, $D_{\rm tr}$
\Ensure $S_{t+1}$, $O_{t+1}$, decision $a_t$, and record $u_t$
\State $\Delta_{\rm sel}^t\gets\Delta_{D_{\rm sel}}(S'_t,S_t)$
\If{$\Delta_{\rm sel}^t\leq0$}
    \State \Return $(S_t,O_t,\textsc{reject},u_t=(\Delta_{\rm sel}^t))$
\EndIf
\State $O'_t\gets$ execute $A$ with $S'_t$ on $D_{\rm tr}$
\State $\Delta_{\rm tr}^t\gets\Delta_{D_{\rm tr}}(S'_t,S_t)$
\State $u_t\gets(\Delta_{\rm sel}^t,\Delta_{\rm tr}^t)$
\If{$\Delta_{\rm tr}^t>0$}
    \State \Return $(S'_t,O'_t,\textsc{accept},u_t)$
\EndIf
\State \Return $(S_t,O_t,\textsc{reject},u_t)$
\end{algorithmic}
\end{algorithm}

Algorithm~\ref{alg:app-update} screens a complete candidate first on the
selection set and then, after a passing first stage, on the training set. Only
an accepted candidate replaces the current skill and supplies the next
training set record. After each round, the loop appends the selected capability, the revision
record, and the screening outcome $(c_t^\star,r_t,u_t)$ to $\mathcal H$. Abstentions and rejected candidates are
retained as revision history, while leaving the current skill unchanged.

\section{Representative Prompt Interfaces}
\label{app:prompt-templates}

SkillFocus issues its model calls through eight structured prompt interfaces
(Table~\ref{tab:prompt-overview}), with runtime records, skill text, evidence,
history, and perspectives substituted by the runner.  We show one interface
per stage; the remaining five follow the same contract.

\begin{table}[t]
\centering
\caption{Overview of the prompt interfaces and their call multiplicity in the
formal runs.}
\label{tab:prompt-overview}
\small
\setlength{\tabcolsep}{3pt}
\begin{tabularx}{\columnwidth}{@{}lXr@{}}
\toprule
Interface & Role & Calls \\
\midrule
\multicolumn{3}{@{}l}{\textit{capability induction}} \\
\texttt{ExtractRequirements} & task $\rightarrow$ requirements & per task \\
\texttt{ProposeCapabilities} & cross-task proposals & 3 \\
\texttt{Consensus} & frozen registry & 1 \\
\texttt{Assign} & requirement $\rightarrow$ capability & batched \\
\addlinespace
\multicolumn{3}{@{}l}{\textit{capability revision}} \\
\texttt{ExploreDiagnoses} & diagnosis hypotheses & 3/round \\
\texttt{IntegrateDiagnosis} & selected diagnosis & 1/round \\
\texttt{ExploreEdits} & edit concepts & 3/round \\
\texttt{IntegrateEdit} & final candidate edit & 1/round \\
\bottomrule
\end{tabularx}
\end{table}

\begin{table}[t]
\centering
\caption{Frozen Explorer perspectives in the formal runs.}
\label{tab:explorer-perspectives}
\small
\setlength{\tabcolsep}{3pt}
\begin{tabularx}{\columnwidth}{@{}>{\raggedright\arraybackslash}p{0.36\columnwidth}>{\raggedright\arraybackslash}X@{}}
\toprule
Stage / perspective & Frozen operational condition \\
\midrule
Diagnosis: missing guidance & No current-skill rule addresses the operation; the explorer leaves \texttt{cited\_rules} empty. \\
Diagnosis: ineffective guidance & An existing rule addresses the operation but fails; the explorer quotes at least one such rule. \\
Diagnosis: behavior organization & Interacting rules create an ordering, conditioning, or precedence problem; the explorer quotes at least two rules. \\
Generation: minimal addition & Propose one non-duplicative standalone rule with \texttt{op=add}. \\
Generation: revise existing guidance & Replace the core content of one existing rule, quoting the rule being revised. \\
Generation: structural change & Change a rule's trigger, ordering, precedence, interaction, or removal target, quoting the affected rule(s). \\
\bottomrule
\end{tabularx}
\end{table}

The runner selects these rows in order, so $\pi^D_j$ and $\pi^G_j$ in
Algorithm~\ref{alg:app-revision} identify the row for Explorer call $j$; the
configured \textit{mechanism-grounded addition} perspective is unused at
$J_G=3$.

\subsection{Capability Proposal}
\label{app:prompt-proposal}
\begingroup
\small
\noindent\hrule
\smallskip
\textbf{Interface:} \texttt{ProposeCapabilities}. Group required behaviors
that recur across distinct tasks and can share skill guidance. Keep each
capability cross-task, observable, and skill-addressable; do not group by
benchmark, tool, operation name, or output artifact. The shared template is
called with one fixed perspective tag $\rho_j$ per independent proposal.
Specifically, $(\rho_1,\rho_2,\rho_3)$ are
\texttt{REQUIREMENT-CENTRIC}, \texttt{INTERVENTION-CENTRIC}, and
\texttt{BOUNDARY-CENTRIC}; the tag changes the grouping lens but not the input
records or output schema. The proposals use only the canonical required-behavior
records and do not use execution outcomes, current skill text, or evaluator
results. Merge behaviors only when the same reusable guidance would normally
improve them; keep them separate when distinct guidance is needed. Do not
optimize for a target number of capabilities, and cite representative behavior
records rather than performing assignment in this call.

\textbf{Output.}
\begin{Verbatim}[fontsize=\small,frame=single,framesep=2pt]
<capability_proposal>
  <capabilities>
    <capability>
      <name>behavioral name</name>
      <definition>
        reusable requirement
      </definition>
      <evidence_refs>
        <ref>RB_ID</ref>
      </evidence_refs>
    </capability>
  </capabilities>
</capability_proposal>
\end{Verbatim}
\noindent\hrule
\endgroup

\subsection{Diagnosis Explorer}
\label{app:prompt-diagnosis-explorer}
\begingroup
\small
\noindent\hrule
\smallskip
\textbf{Interface:} \texttt{ExploreDiagnoses}. Given the selected capability,
evidence bundle, and current skill, develop one evidence-supported hypothesis
connecting an observed execution pattern to an editable skill location. One
Explorer returns one hypothesis; it does not write an edit. The Explorer cites
the evidence records and any current-skill rules required by its configured
perspective, and abstains when the bundle cannot support one defensible
hypothesis.

\textbf{Output.}
\begin{Verbatim}[fontsize=\small,frame=single,framesep=2pt]
<diagnosis_explorer_result>
  <status>
    supported|insufficient_evidence
  </status>
  <hypothesis>
    one evidence-grounded diagnosis
  </hypothesis>
  <cited_rules>
    <rule>verbatim current-Skill rule</rule>
  </cited_rules>
  <evidence_refs>
    <ref>EVIDENCE_ID</ref>
  </evidence_refs>
</diagnosis_explorer_result>
\end{Verbatim}
\noindent\hrule
\endgroup

\subsection{Edit Integrator}
\label{app:prompt-edit-integrator}
\begingroup
\small
\noindent\hrule
\smallskip
\textbf{Interface:} \texttt{IntegrateEdit}. Select or synthesize the smallest
coherent edit from the proposed concepts and the diagnosis. Return
\texttt{no\_valid\_edit} if no operation has a precise target and a reusable
content change. The returned candidate records its operation, target contract,
preservation scope, trigger/action/check, expected change, falsifier, and
history assessment in addition to the target-facing content. The integrator
also checks the formal rejection memo, preserves effective unrelated skill
content, avoids benchmark-specific hard-coding, and requires a concrete worked
example in the target-facing edit.

\textbf{Output.}
\begin{Verbatim}[fontsize=\small,frame=single,framesep=2pt]
<candidate_edit_result>
  <status>
    candidate|no_valid_edit
  </status>
  <op>add|replace|remove</op>
  <target>
    exact target or insertion location
  </target>
  <trigger>
    when the rule applies
  </trigger>
  <action>
    what the agent should do
  </action>
  <check>
    what the agent should verify
  </check>
  <preserve>
    effective behavior to preserve
  </preserve>
  <expected_change>
    failure behavior expected to change
  </expected_change>
  <content><![CDATA[
exact target-facing Skill edit including
the worked example
]]></content>
</candidate_edit_result>
\end{Verbatim}
Audit fields for the target specification, preservation scope, history
assessment, falsifier, and rationale are omitted.
\noindent\hrule
\endgroup

\section{Use of Large Language Models}
\label{app:llm-usage}

Large language models are components of the studied system: GPT-5.4 is the
optimizer model for capability induction, diagnosis, and edit generation, and
DeepSeek-V4-Flash is the fixed target model that executes benchmark tasks; the
baselines use the same models in these roles.

\end{document}